\documentclass[sigconf]{acmart}
\copyrightyear{2024}
\acmYear{2024}
\setcopyright{rightsretained}
\acmConference[WWW '24]{Proceedings of the ACM Web Conference 2024}{May 13--17, 2024}{Singapore, Singapore}
\acmBooktitle{Proceedings of the ACM Web Conference 2024 (WWW '24), May 13--17, 2024, Singapore, Singapore}
\acmISBN{979-8-4007-0171-9/24/05}
\acmDOI{10.1145/3589334.3645631}

\makeatletter
\gdef\@copyrightpermission{
 \begin{minipage}{0.3\columnwidth}
  \href{https://creativecommons.org/licenses/by/4.0/}{\includegraphics[width=0.90\textwidth]{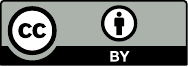}}
 \end{minipage}\hfill
 \begin{minipage}{0.7\columnwidth}
  \href{https://creativecommons.org/licenses/by/4.0/}{This work is licensed under a Creative Commons Attribution International 4.0 License.}
 \end{minipage}
 \vspace{5pt}
}
\makeatother

\usepackage{amsmath}
\usepackage[inline]{enumitem}
\usepackage{ wasysym }

\usepackage{balance}

\usepackage{listings}
\newcommand*{\tc}[1]{\textbf{#1}}

\begin{document}

%%
%% The "title" command has an optional parameter,
%% allowing the author to define a "short title" to be used in page headers.
\title{Perennial Semantic Data Terms of Use for Decentralized Web}

%%
%% The "author" command and its associated commands are used to define
%% the authors and their affiliations.
%% Of note is the shared affiliation of the first two authors, and the
%% "authornote" and "authornotemark" commands
%% used to denote shared contribution to the research.
\author{Rui Zhao}
\orcid{0000-0003-2993-2023}
\affiliation{%
  \institution{University of Oxford}
  \city{Oxford}
  \country{UK}
}
\email{rui.zhao@cs.ox.ac.uk}

\author{Jun Zhao}
\orcid{0000-0001-6935-9028}
\affiliation{%
  \institution{University of Oxford}
  \city{Oxford}
  \country{UK}}
\email{jun.zhao@cs.ox.ac.uk}

%%
%% By default, the full list of authors will be used in the page
%% headers. Often, this list is too long, and will overlap
%% other information printed in the page headers. This command allows
%% the author to define a more concise list
%% of authors' names for this purpose.

\begin{abstract}
In today's digital landscape, the Web has become increasingly centralized, raising concerns about user privacy violations. Decentralized Web architectures, such as Solid, offer a promising solution by empowering users with better control over their data in their personal `Pods'.
However, a significant challenge remains: users must navigate numerous applications to decide which application can be trusted with access to their data Pods. This often involves reading lengthy and complex Terms of Use agreements, a process that users often find daunting or simply ignore. This compromises user autonomy and impedes detection of data misuse. We propose a novel \emph{formal} description of Data Terms of Use (DToU), along with a DToU reasoner.
Users and applications specify their own parts of the DToU policy with local knowledge, covering permissions, requirements, prohibitions and obligations. Automated reasoning verifies compliance, and also derives policies for output data. This constitutes a ``perennial'' DToU language, where the policy authoring only occurs once, and we can conduct ongoing automated checks across users, applications and activity cycles.
Our solution is built on Turtle, Notation 3 and RDF Surfaces, for the language and the reasoning engine. It ensures seamless integration with other semantic tools for enhanced interoperability. We have successfully integrated this language into the Solid framework, and conducted performance benchmark.
We believe this work demonstrates a practicality of a perennial DToU language and the potential of a paradigm shift to how users interact with data and applications in a decentralized Web, offering both improved privacy and usability.
\end{abstract}

%%
%% The code below is generated by the tool at http://dl.acm.org/ccs.cfm.
%% Please copy and paste the code instead of the example below.
%%



%%
%% This command processes the author and affiliation and title
%% information and builds the first part of the formatted document.
\maketitle

\section{Introduction}

After years of development, the Web has become an indispensable part of people's life. However, the centralization of the Web has risen as a pressing challenge, leading to various issues like pervasive user behavioural manipulation, privacy breaches and an imbalance of power \cite{stein_you_2023,zuboff2019age}. Decentralization is viewed as a potential solution to address these issues \cite{meurisch_privacy-preserving_2020,ion_home_2011}, and initiatives like Solid (Social Linked Data) \cite{sambra_solid_2016} have gained attention for their aim to return \textit{data and control} to the user, while respecting the openesss and fairness of Web standards and infrastructures.

In a decentralized Web, users store their data in their own storage (such as Solid Pods), and applications must request permission to use and store data there. This shift in data control has the potential to reduce `vendor-lock-in', as it limits the privileged ownership of data currently prevalently observed in large platforms. Furthermore, it is also crucial to fostering competition among different applications. However, one aspect that has received less attention in the decentralized setting is \textit{how users can sensibly decide} which application should be granted permission to access their data.

Assume in the decentralized setting, Alice wants to use a shopping app to buy shoes, which may require access to several types of data in her Pod, including shoe size, delivery address, and billing information. She is concerned about how such an app may handle her data ethically. In the meantime, Bob, the developer of a shopping app, HappyShop, wants to build trust with users and is willing to provide descriptions on how HappyShop handles users data through its `Terms of Use'. However, users like Alice typically do not read these terms (aka.~``the biggest lie on the Internet'') because of information overload and their length~\cite{mcdonald_cost_2008,obar_biggest_2020}. The problem is exacerbated in the decentralized setting with its numerous apps that may require users' decision regarding access to their data Pods, especially if, e.g.~an accounting app, TotalAcc, reuses the data produced by HappyShop (e.g.~order details). As a result, users autonomy may still be compromised in the decentralized Web, and can lead to many issues related to data misuse~\cite{zimmeck_automated_2016,tesfay_i_2018,breaux_eddy_2014} or loss of trust~\cite{huh-yoo_its_2020}.

Facing these challenges, we propose the concept of ``perennial'' Data Terms of Use (DToU), characterizing a formal language model that addresses the following \textbf{challenges} in a decentralized Web context: \textbf{C1}) expressing data provider's DToU for their data; \textbf{C2}) imparting application's (developer's) DToU on how they handle data; \textbf{C3}) performing compliance checking over data usage requests; \textbf{C4}) supporting DToU policy reusing across applications and data providers;
\textbf{C5}) facilitating apt DToU-compliant cross-application data sharing.
With such a language model, automated reasoning can be performed thus only exposing distilled important information to users, reducing the amount of information and numbers of decisions exposed to the user, thus incentivizing responsible handling of Terms of Use.
It is called ``perennial'' because a stakeholder only needs to specify the DToU once in the beginning, and it can be reused across activity cycles and across stakeholders, just like the plants being implanted once and kept growing in the future.

The \emph{perennial} concept is in contrast to \emph{sticky policy} \cite{mont_towards_2003,pearson_sticky_2011} which proposed principles for supporting distributed DToU compliance, but, in the core definition, only explicitly discussed requirements 1 and 3. \emph{Sticky policy} supports DToU-compliant cross-application data sharing, but suggests the policy staying the same regardless of data processing history, thus is not \emph{apt}, leading to the potential of frequent user disruption.
Existing research on policy languages, often coined as access control~\cite{qiu_survey_2020} and usage control~\cite{sandhu_usage_2003,lazouski_usage_2010}, also proposed formal models for expressing certain parts of DToU. There are different flavours of them, targeting at different scenarios, with diverse properties and capacities. As we will introduce in Sec \ref{sec:related_research}, they provide many design principles and concepts that are useful across contexts, but impose their individual limitations.

In this paper, we propose a \emph{perennial} policy language that addresses challenges from decentralization while maintaining expressiveness, by using heterogeneous yet interoperable data policy and application policy. The language and reasoning mechanism are built on semantic technologies, namely Turtle \cite{beckett_rdf_2014}, Notation 3 \cite{berners-lee_n3logic_2008} and RDF Surfaces \cite{hochstenbach_rdf_2023}. This facilitates the creation of a common vocabulary, and enables integration with other semantic tools, particularly ontologies and ontological reasoning. Furthermore, our approach is integrated with Solid, a decentralized user-focused Web architecture, and we evaluate its performance across various workloads\footnote{See our repo \url{https://github.com/OxfordHCC/solid-dtou} for the source code. Experimental logs and other information are in the supplementary artifact at https://doi.org/10.5281/zenodo.10685603.}.
To the best of our knowledge, we are the first work proposing a (semantic) perennial policy language that supports stakeholders expressing their DToU in the context of the decentralized Web.
We believe this work provides a good starting of the paradigm shift for enhancing user autonomy and control.

\section{Related Research}
\label{sec:related_research}
\begin{table*}[ht]
    \centering
    \caption{Summarization of features under different categories of related policy languages.}
    {\footnotesize  C1 - C5 refers to the Challenges 1 - 5 identified for perennial language. For column C1, A means authorization, O means obligation, and + means additional features. For columns C3 \& C4, \checked~means true, \CheckedBox~means true if using same environmental information schema. For columns C2, C5 and Condition, A means application, C means capacity, D means data type / category, E means environmental information, \textit{\textbf{E}} means entire environmental information exposed in knowledge base, I means multi-input, M means use mode, O means multi-output, P means purpose, T means transformation, U means user, X means external information.}
    \label{tab:related_research_features}
    
    \begin{tabular}{ccccccccc}
        \toprule
        Language & C1 & C2 & C3 & C4 & C5 & Condition & Policy Author & Format \\
        \midrule
        E-P3P\cite{karjoth_platform_2002} & AO &  & \checked & \CheckedBox &  & EPU & Supervisor & Custom \\
        XACML\cite{noauthor_extensible_2013}/ODRL\cite{noauthor_odrl_2018} & AO+ &  & \checked & \CheckedBox & & EPU & Owner & XML/JSON-LD \\
        WAC\cite{noauthor_web_2022}/ACP\cite{noauthor_access_2022} & A & & \checked & & & AU/AUX & Owner & Turtle \\
        P2U\cite{iyilade_p2u_2014} & A & & \checked & & & AU & Owner & XML \\
        LaBAC\cite{biswas_label-based_2016} & A & & \checked & & & AU & Supervisor & Custom \\
        AIR\cite{kagal_using_2008} & A & & \checked & \CheckedBox & & \textit{\textbf{E}}U & Owner & N3 \\
        Thoth\cite{elnikety_thoth_2016} & A & & \checked & & & EUX & Owner & Logic-like \\
        Eddy\cite{breaux_eddy_2014} & AO & MP &  & \CheckedBox & T & MPU & Supervisor & OWL-DL  \\
        LoNet\cite{havard_d_johansen_enforcing_2015} & A & & \checked & \checked & T & EU & Supervisor & Custom \\
        CamFlow\cite{pasquier_camflow_2017} & A & C & \checked & \checked & T & C & Owner & Custom \\
        Smart object\cite{sagirlar_decentralizing_2018} & A & DIP & \checked & \checked & T & DP & Owner & Custom \\
        Dr.Aid\cite{rui_zhao_draid_2021} & AO & IP & \checked & \checked & OT & EPU & Owner & Custom 
        \vspace{0.3em} \\
        \textbf{This paper} & AO & CDIP & \checked & \checked & OT & CEPU & Owner & Turtle \\
        \bottomrule
    \end{tabular}
\end{table*}

Several explorations have delved into the utilization of computer-interpretable formal encoding of policies to enable (semi-)automated decision-making of data usage authorization. These range from the classical access control to more advanced dynamic usage descriptions. In this section, we examine this body of research and discusses their relevance to a decentralized Web context. The main features of several closely-related policy languages are summarized in Table \ref{tab:related_research_features} (see Appendix \ref{sec:appendix:related-work-extension} for further details of term explanation), and this section discusses their general properties. Where relevant, we will refer to the challenges or concepts shown in the table.

The most well-known line of research involves various access control models, each based on different principles. For example, models such as Mandatory Access Control (MAC) \cite{sandhu_access_1994}, Access Control List (ACL), Role-Based Access Control (RBAC) \cite{sandhu_role-based_1996} or Attribute-Based Access Control (ABAC), can be based on information ranging from narrower details like user identity to broader categories such as user groups, and contextual information. Several languages uses or implements them, such as E-P3P \cite{karjoth_platform_2002}, WAC \cite{noauthor_web_2022} and P2U \cite{iyilade_p2u_2014}. There is also research that falls in the middle ground, such as Label-Based Access Control (LaBAC) \cite{biswas_label-based_2016}, a simplified variant of ABAC while more expressive than MAC and RBAC.

Other policy languages may also use some concepts from access control models, with additional features, such as eXtensible Access Control Markup Language (XACML) \cite{noauthor_extensible_2013}, a widely-known XML-based standard for ABAC with additional constructs like obligations for cloud services, and Open Digital Rights Language (ODRL) \cite{noauthor_odrl_2018}, an expressive policy language serializable to JSON-LD (also to be discussed later).
Thoth \cite{elnikety_thoth_2016}, on the other hand, uses its own logic-like policy language to express policies not only for read action, but also for write, update and declassification actions. It permits intricate evaluations of formulae that involve rich operators and the incorporation of external sources of information during policy evaluation.

These approaches vary in terms of expressiveness and usage contexts.
However, they tackle the problem either purely relying on the data providers, or from a holistic view requiring a `supervisor'. In both settings, they require the policy author with abundant knowledge to actively compile and maintain policies during personnel changes (e.g.~adding a new application, or assigning a user/operator with a new role).
% However, they usually rely on a finite/closed world, where it is possible to enumerate or pre-identify all data consumers (humans or applications), and/or a `supervisor' with abundant knowledge will actively compile and maintain policies during personnel changes (e.g.~adding a new application, or assigning a user/operator with a new role). Without doing so, the policy will not be able to cope with changes.
While this assumption is reasonable for managing data usage at an institutional level, it presents challenges in contexts with multiple independent stakeholders that engage and disengage dynamically (esp.~\tc{C2}, \tc{C4} and \tc{C5}), as is the case in a decentralized Web.
For example, they can only define the policy \emph{after} a user like Alice has determined whether a shopping app like HappyShop should be granted permission or not, rather than supporting proactive decisions.

In contrast, some research recognizes the dynamic nature of data and applications, and offers different solutions. LoNet \cite{havard_d_johansen_enforcing_2015}, for example, employs Information Flow Control (IFC) for RBAC, along with conditional predefined policy evolution (\tc{C5}). It tries to address the expressiveness through the so-called meta-code (an external script) for checking additional policy information like time. However, the meta-code is arbitrary code and difficult to statically verify.
CamFlow \cite{pasquier_camflow_2017}, on the other hand, borrows concepts from Decentralized IFC (DIFC) \cite{myers_decentralized_1997}, and employs homogeneous policy constructs (tags and labels) to encode both the data policy and application capability, and checks their compatibility (\tc{C2} \& \tc{C4}). It expresses policy evolution (\tc{C5}) by adding or removing tags for output (policy) based on input (policy). 
Smart object \cite{sagirlar_decentralizing_2018} uses complex constructs to define permitted and prohibited use of data, and combines this with the application information (using relational calculus) (\tc{C2}) to derive policies for output data (\tc{C5}), assuming and leveraging a tabular structure of data.
Dr.Aid \cite{rui_zhao_draid_2021} focuses on the context of data-intensive scientific workflows, employing different structures to express data rules and process rules (\tc{C2}). It supports policy reasoning and derivation for workflow graphs composed of multi-input-multi-output processes (\tc{C5}).
A common feature among these approaches is the separation of data policy from application policy.
% , using pre-defined reasoning mechanisms to facilitate compliance checking.
They have demonstrated that this separation allows the reasoner to verify if an application complies with the data policy, aligning with our intended goal. However, they often have relatively limited expressiveness compared to the earlier research.

There are also policy languages that utilize semantic technologies or are tailored for specific use cases within the decentralized Web. One notable example is ODRL~\cite{noauthor_odrl_2018}, which offers a comprehensive set of concepts, including permissions, prohibitions, obligations, remedies, conditions, purposes and agents. However, originated as a data right expression language, ODRL primarily focuses on expressing what is (not) permitted for the data, and lacks a corresponding mechanism for expressing application information. While efforts are being made to address this issue \cite{esteves_odrl_2021}, there is still uncertainty regarding the eventual resolution.
% the representation of much information is context-dependent, affecting the generality of the reasoner (validator).
The AIR \cite{kagal_using_2008} policy language, based on Notation 3 (N3) \cite{berners-lee_n3logic_2008}, specifies permitted and prohibited actions for data processing. It allows for the expression of custom rules directly on the contextual knowledge base of data processing. However, in the absence of general agreements across applications, this requires the policy author to have the knowledge of information and structure of the knowledge base, resulting in a strong coupling with the application.
Eddy \cite{breaux_eddy_2014} examined real-life Terms of Use from online services and proposed a policy language based on OWL-DL to express key data requirements, including permissions, obligations and prohibitions. The language distinguishes between different use modes (collect, use, retain and transfer) and demonstrated compliance checking between two policy sets of two services. However, there is no clear demonstration of how data providers can utilize this language, as it necessitates highly detailed descriptions.
Solid, a decentralized Web architecture based on Linked Data, has its own policy languages, mainly for access control purposes. This includes WAC \cite{noauthor_web_2022}, a policy language for expressing ACL, and ACP \cite{noauthor_access_2022}, an advanced policy language that supports a broad range of conditions, such as Verifiable Credentials. WAC and ACP leverage contextual information exposed by Solid but are less expressive compared to ODRL and AIR.
% In general, these languages cannot be directly applied to support data usage expressions in the decentralized setting we target at due to their individual limitations. %In addition, there is no clear solution for associating policies with output data of the application, leading to an undesired impact for the interaction across applications if they share data.Reusing them would introduce heavy overload when designing a new language, though their expressiveness may be useful in the future when extending the language.

In summary, the various policy languages exhibit different features tailored for their specific use cases. Among them, we find the second category of research, which separates data policy from application policy, to be the most suitable for our intended design context.
However, they are not directly applicable to our context, primarily due to limitations in their expressiveness.
% For example, the tags of CamFlow are rudimentary and struggle to convey contextual information effectively. Smart Object enforces a rigid tabular structure of data, and offers limited support for conditions based on data category and usage purposes. Dr.Aid,  on the other hand,  primarily supports obligations and simulated prohibitions, with only limited expressions related to applications.
Our research takes inspiration from them, addresses such limitations, and provides better expressiveness in an extensible way.

\section{A perennial policy language}

\subsection{Language design}
Broadly speaking, our language model consists two parts: the \emph{data policy} and the \emph{app(lication) policy}. This design is crucial to enable automated data access negotiation between a data owner and a data consumer (e.g. applications).
The \textit{data} policy enables data providers to define policy-related metadata and expectations for the data consumers. The \textit{app} policy allows app developers to encode the promises and expectations for accessing the data by the application. 

\begin{figure*}
    \centering
    \includegraphics[width=.95\textwidth]{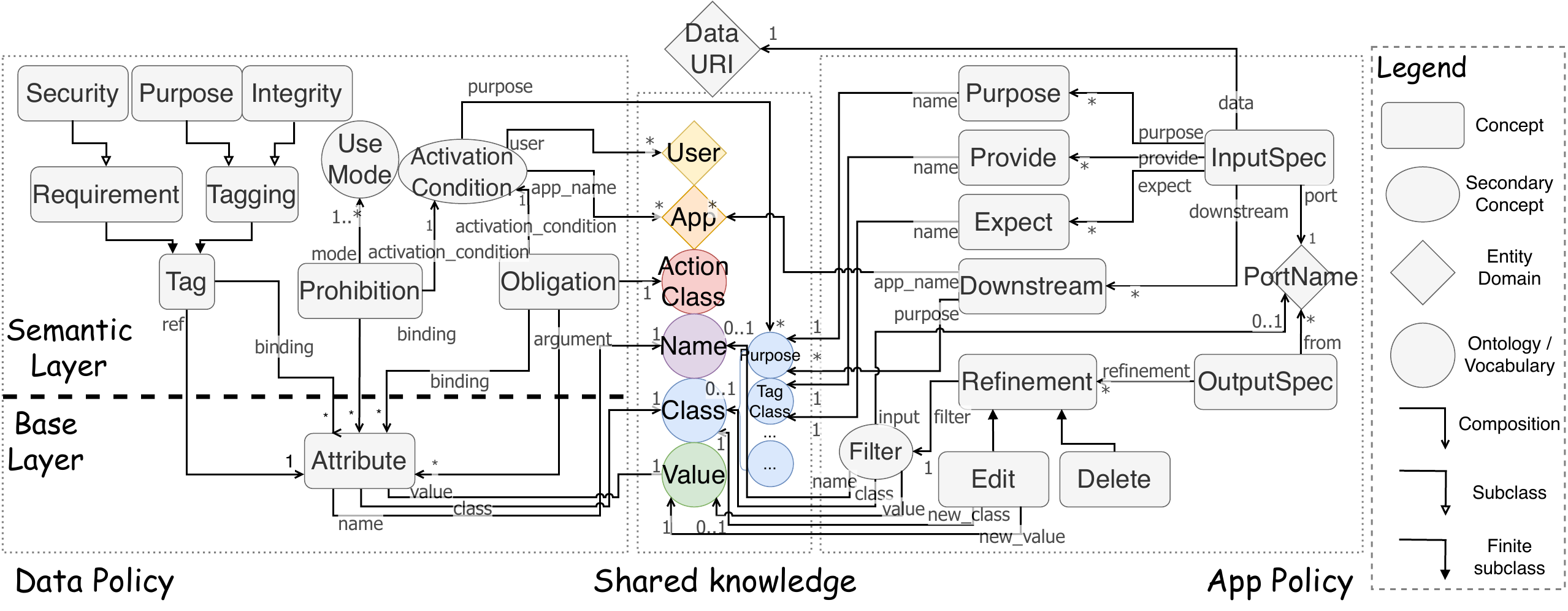}
    \caption{Language design and relation between concepts}
    \label{fig:lang-design}
\end{figure*}

The reasoner performs three types of tasks: a) conformance check: deciding whether the application can use the data; b) obligation check, assessing what obligations are activated by the application; and c) policy derivation: determining the policy for output data and saving a data owner to define data usage policy for derived data. Figure \ref{fig:lang-design} gives an overview of the language and the relation between different concepts.

In the following, we provide more detailed descriptions about the language, with all examples expressed in Turtle\footnote{For simplicity, we omit the prefixes, and only use \lstinline|:| for named nodes.} \cite{beckett_rdf_2014}. 

\subsubsection{Data Policy}

Conceptually, the data policy consists of two layers: attributes and semantics. Attributes form the base layer, expressing \emph{information} about the data and/or the policy. On top of that is the semantic layer, providing semantic concepts like tags, prohibitions and obligations, while referencing information expressed in the attribute layer. In the Turtle syntax of the policy encoding, each statement/tuple is expressed as a node with corresponding type and properties.

More precisely, each \emph{attribute} is a simple tuple: \lstinline|(name, class, value)|. While they alone do not carry semantics for the policy, they are designed with a flexible structure to describe information about data and policy. For example, attributes can be used to define Alice as an author, e.g.~\lstinline|(:author, :string, "Alice")|, or encode a textual description (e.g.~\lstinline|(:ack-text, :string, "This dataset is by Alice")|), or specify the data fields (e.g.~\lstinline|(:col-2, :field, :column-2)|).
This concept of \emph{attributes} is borrowed from Dr.Aid \cite{rui_zhao_draid_2021}, which has been demonstrated as a powerful structure for supporting policy derivation. %Essentially. they are similar to the attributes in ABAC, but are information contained within the policy, instead of general information retrievable from third-party sources through Policy Information Point.
The following example shows how Alice defines her email address information as an attribute -- the attribute is a kind of `string' and has the value of ``alice@a.b''.
\begin{lstlisting}
:attr1 a :Attribute;
  :name :alice-email;
  :class :string;
  :value "alice@a.b".
\end{lstlisting}
\emph{Tags}, from the semantic layer of our model, can be used to specify the requirements or available resources. Two typical categories of tags are \emph{security} and \emph{integrity}: \emph{security} specifies which security clearance an application needs to possess to use the data; and \emph{integrity} identifies the integrity level(s) this dataset has, to be requested by the application (policy). %Therefore, \emph{security} tags of the data policy should be a \emph{subset} of that of the app policy, while \emph{integrity} tags should be the \emph{superset}.
Inspired by CamFlow \cite{pasquier_camflow_2017} (see Sec \ref{sec:related_research}), tags can be used to facilitate \emph{capacity} of conformance check, see later Sec \ref{sec:lang:reasoning:conformance-check}. %of the application policy should be compatible with the tags of the data policy to permit data usage, by checking the superset and subset relation among the sets of tags of the same type, as to be introduced later in .

The descriptor of a tag (i.e.~which tag it is) is associated with the \lstinline|class| of an \emph{attribute} using \lstinline|attribute_ref|; other information in the \emph{attribute} is not used by a tag.
Our policy language also allows expressing additional categories of tags, such as \emph{Purposes}.

For example, the following excerpt (from Appendix \ref{sec:appendix:policy:data:payment-info}) shows how \emph{tags} can be used to define that Alice requires any application to respect \lstinline|banking| security level (to use payment info):
\begin{lstlisting}
:tag2 a :SecurityTag;
    :attribute_ref :attr-tag2;
    :validity_binding :attr2.

:attr-tag2 a :Attribute;
    :name :tag-2;
    :class :banking;
    :value :nil.

:attr2 a :Attribute;
    :name :det;
    :class :data-content;
    :value :payment-details.
\end{lstlisting}
This example also contains a \emph{validity binding} for that tag,  meaning that the validity of the tag is dependent on the existence of the referenced attribute(s), i.e.~\lstinline|:attr2|, denoting the exact content of \lstinline|:payment-details|.

\emph{Prohibitions} specify additional restrictions on the data consumer, independent of the capacity (i.e.~\emph{tags}) of the application. A prohibition contains an \emph{activation condition}, which is a matcher against the context, e.g.~user, application and purpose. If the condition matches the usage context, the usage is deemed prohibited. %All fields can be optional, i.e.~matching any value, or have multiple values, i.e.~matching one of the values. 
\emph{Validity bindings} can also be specified.
Additionally, our model also offers an extension point, allowing users to specify the \lstinline|:mode| when a match is considered. Currently, we only support the \lstinline|:Use| mode, denoting the reading or processing of the data. We plan to explore other types of use modes to be integrated.

For example, Alice dislikes a payment processor, \lstinline|<http://duckpay.com/>|, and does not want it to use her payment info, either directly or indirectly. This can be encoded as:
\begin{lstlisting}
:pr1 a :Prohibition;
    :mode :Use;
    :activation_condition
        [ :app_name <http://duckpay.com/> ];
    :validity_binding :attr2.
\end{lstlisting}

The following example for shoe size shows that Alice expects any application to email her when used for research purposes:
\begin{lstlisting}
:ob1 a :Obligation;
    :obligation_class :send-email;
    :args (:attr1);
    :activation_condition [ :purpose :research ].
\end{lstlisting}

\paragraph{Policy set}
Finally, the policy terms explained above should be put together as a policy set. A policy set contains a \lstinline|Policy| node which contains the relevant policy terms, and a \lstinline|Data| node which pairs the policy and the data IRI that this policy applies to. For example, the policy set for Alice's payment information looks like:
\begin{lstlisting}
:data-payment a :Data;
    :uri <http://a.b/payment-info>;
    :policy :policy-1.

:policy-1 a :Policy;
    :attribute :attr-tag2, :attr-tag3,:attr-tag4, :attr2;
    :security :tag2;
    :purpose :tag3, :tag4;
    :prohibition :pr1.   
\end{lstlisting}

\subsubsection{Application Policy}

An \emph{app policy} contains basic information about the application, and the policy specification for the inputs and outputs. In addition, it also specifies relevant \emph{downstream} data consumers (e.g.~third-party APIs to send data to) that this application will use.

An application may take multiple inputs, and thus multiple \emph{input specifications} (\lstinline|:InputSpec|). Normally, each input specification describes basic information about that input (\emph{name} and \emph{data}) and its \emph{capacity}: the \emph{tags} it conforms to, including the \emph{security} levels, the \emph{integrity} it expects, and the \emph{purpose} it will use the (input) data for.
If the application sends the input data to an external location for processing (e.g.~an API call), a \emph{downstream} should be specified as a simplified app policy for that downstream stakeholder, specifying the (app) \emph{name}, \emph{user} and \emph{purpose} of that \emph{downstream}\footnote{The app developer should verify that the downstream capacity is in line with that of the input specification, so its tags do not need to be explicitly expressed.}.

For example, HappyShop states this input specification for reading the payment information:
\begin{lstlisting}
:input1 a :InputSpec;
    :data <http://a.b/payment-info>;
    :port [:name "payment-info-in"];
    :security :banking;
    :purpose :making-payment;
    :downstream [
        :app_name <http://goodpay.com/>;
        :purpose :making-payment ].
\end{lstlisting}
It says an input port named \lstinline|"payment-info-in"| reads data from \lstinline|<http://a.b/payment-info>|, and promises to comply with security level \lstinline|:banking|, and will use the data only for purpose of \lstinline|:making-payment|; it will send the data to a downstream, named \lstinline|<http://goodpay.com/>|, for purpose of \lstinline|:making-payment|.
It is compatible with Alice's data policy for payment info.

Similarly, an application may wish to store data into user's Pods, so it may need to specify multiple \emph{output specifications}. Apart from the \emph{name}, each \emph{output specification} describes the related input that this output data is derived \emph{from}, and the \emph{refinements} that the data policies are subject to.

The \emph{from} statement, (during reasoning) associates the output with a set of data policies, each pertinent to the input used to derive the output. In a higher level, this reflects the general information flow in the application.
The removal or change of information is captured by a \emph{refinement}, which expresses how (the \emph{attributes} of) the data policies should be modified to reflect the processing that has been applied to the data. Two types of \emph{refinements} are supported: \emph{delete} and \emph{edit}. A \emph{delete} has a \emph{filter}, meaning that all attributes matching the \emph{filter} should be deleted, and the \emph{filter} is used to specify the matching \emph{attribute} information: \emph{name}, \emph{class} and \emph{value}.
Similarly, an \emph{edit} means that any \emph{attribute} matching the filter will be assigned a new class and value. %Therefore, it has a \emph{filter}, and a new \emph{class} and \emph{value} for the updated \emph{attribute}.

The following example output specification shows how HappyShop may record purchase histories in user's Pod, which contains derived data (the copy) of the delivery address, and a declassified version of payment details:
\begin{lstlisting}
:out1 a :OutputSpec;
    :from [ :name "address-in" ],
        [ :name "payment-info-in" ];
    :refinement :refine-no-payment-details.

:refine-no-payment-details a :Delete;
    :filter [
        :class :data-content;
        :value :payment-details ].
\end{lstlisting}
This policy snippet means this output is related to the data from the inputs named \lstinline|"address-in"| and \lstinline|"payment-info-in"|, and has one refinement, which will delete all attributes of class \lstinline|:data-content| and value \lstinline|:payment-details|, reflecting a fact that the output data will not contain payment details even though it uses payment data.

The \emph{refinements} directly operate on \emph{attributes}, but references to attributes (\emph{bindings}) in the semantic layer ensure that all operations will be inferred to related semantic concepts too. For instance, if an \emph{attribute} is deleted, any \emph{tags}, \emph{prohibitions} and \emph{obligations} that have a \emph{binding} to this \emph{attribute} will also be deleted.

Similar to data policy, the app policy statements should be put together as a policy set. Please refer to Appendix \ref{sec:appendix:policy:app:policy-set} for example.

\subsubsection{Shared vocabulary}
It is worth noting that shared vocabularies are assumed in all related research, with different levels of difficulties to achieve. In our work, this is achieved by making concepts explicit and using URIs/IRIs. This allows easy and decentralized provision of them, such as using OWL ontologies \cite{w3c_owl_working_group_owl_2012} as vocabularies.
There are five sorts of vocabularies to be shared, as seen in the middle of Figure \ref{fig:lang-design}. They are centred around the data provider's wills, and thus is most natural to be provided by them, or some intermediaries.
We mainly identify this mechanism, and leave this to the practitioners to consolidate the vocabularies. Relatedly, OWL reasoning may be integrated and performed before policy reasoning to maximize interoperability.

\subsection{Reasoning}

Our language accommodates three types of reasoning tasks: conformance check, obligation check, and policy derivation.
This section explains the reasoning mechanism in more details.

In general, the reasoning rules can be expressed using first-order logic, encoded using RDF Surfaces \cite{hochstenbach_rdf_2023} and Notation 3 (N3) \cite{berners-lee_n3logic_2008} in our implementation. N3 is a language supporting the expression of both semantic data and reasoning rules with a rich syntax; RDF Surfaces, building on N3, provides an (easy) translation of first-order logic (FOL), including representing negations.
Some of these rules contain explicit negations, which are implemented using N3 built-ins like \lstinline|log:collectAllIn|, enabling scoped negation-as-failure (SNAF) \cite{berners-lee_n3logic_2008}.
For the sake of brevity, we only introduce the foundational principles here, and refer the reader to Appendix \ref{sec:appendix:axiom} for the axioms.

\subsubsection{Context preparation}

Before embarking on the three reasoning tasks, it is essential to inject the contextual information and link the application policy with data policy.

The contextual information (see e.g.~Appendix \ref{sec:appendix:policy:usage-context}) specifies the relevant app policy $app\_pol$, the $user$ and the $time$ of data usage. These parameters are only known during actual application requests and should be provided to the reasoner dynamically each time. Each reasoning task involves a distinct $UsageContext$.

Furthermore, all relevant policy content is added to the same knowledge base. This leverages the fact that Turtle is a sub-language of N3, making all policy specifications in Turtle valid N3 statements. The linkage between data policy and app policy is established by identifying the data URIs as specified in their respective fields. This is achieved through our reasoning rules, removing the need for additional injection of data policy into the corresponding inputs.

\subsubsection{Conformance check}
\label{sec:lang:reasoning:conformance-check}

The fundamental purpose of our policy language is to determine whether a data usage should be permitted, i.e.~conformance checking. In our language, there are three types of conflicts to be checked: unsatisfied requirements, unmatched expectations, and prohibited uses. (See Appendix \ref{sec:appendix:axiom:conformance-check}.)

An \emph{unsatisfied requirement} occurs when a \emph{requirement tag} (e.g.~security) in the data policy is absent in the app policy. Therefore, the reasoning process involves determining all corresponding inputs and data, and verifying their requirement tags.

Conversely, an \emph{unmatched expectation} arises when a \emph{tag expectation} (e.g.~integrity) in the app policy is missing in the data policy. This task also covers the verification of whether all purposes are permitted.
Both the reasoning about \emph{unsatisfied requirement} and \emph{unmatched expectation} require a 'closed-world assumption,' as it is essential to determine if a tag does not exist. Because the policy documents are as the sole reliable source of information, we utilize SNAF provided by N3 built-ins.

A \emph{prohibited use} is identified when a prohibition is triggered. Prohibitions have their own semantics, including activation conditions, and should be checked in accordance with these conditions.%Unlike the tags, a prohibition has its own semantics, i.e.~the activation conditions, and should be checked in accordance with that.

\subsubsection{Obligation check}

Our policy language can also reason about the obligations triggered during data use. This process is similar to checking prohibitions and involves verifying the  \emph{activation conditions}. But because obligations contain arguments that are references to attributes, values of these attributes also need to be returned from the query for further assessment in application logic. (See Appendix \ref{sec:appendix:axiom:obligation-check}.)

\subsubsection{Policy derivation}

When the application produces output data (i.e.~storing data to users' Pods), policy derivation becomes crucial to produce derived data policy for the output, based on the output specification. This aspect is central to (addressing challenge 5 of) the \emph{perennial} nature of our language. Policy derivation involves merging the data policies from all corresponding \lstinline|from| inputs and performing \emph{refinements}, for each corresponding output port. (See Appendix \ref{sec:appendix:axiom:policy-derivation}.)

The derivation of \emph{output attributes} is a primary focus because these attributes are vital for handling output policies for the semantics layer, particularly for bindings.
Because of \emph{refinements}, each input attribute will either have a copy, be edited, or cease to exist in the output. This process creates new nodes for the output attributes in the RDF graph, corresponding to the existential quantifier in the conclusions.
The linkage between the input and output attributes is also recorded,
which will be used for policy derivation of the semantic layer.
SNAF plays a crucial role here as it determines what `happens' to the rest of the input attributes that do not have a matching \emph{refinement} -- there should be an output attribute with identical name, class and value.

The \emph{tags} in an output are based on the collection of all tags of related inputs, while removing those tags with deleted attribute bindings. 
%The output tags will have \lstinline|attribute_ref| and \lstinline|validity_binding| to the corresponding output attributes.
Because tags have categories, they can be treated uniformly when reasoning about their existence, and the reasoner can make use of the categories afterwards.

The \emph{output obligations} and \emph{output prohibitions} are derived similarly. We first check if any binding is deleted, like for \emph{output tags}.
If not, a new node for obligation (or prohibition) is created, replicating all fields of the original obligation (or prohibition) in input. For obligations, that covers the obligation definition (obligated action class and arguments), validity binding and activation condition; for prohibitions, that covers the use mode, validity binding and activation condition.

Intuitively, the attribute references (of statements in the semantics layer) for the output policies will be the corresponding output attributes, instead of the input attributes.

\section{Solid Integration}

To test and demonstrate the language, we integrated the language into Solid, a decentralized Web architecture based on Linked Data that emphasizes on user autonomy. This allowed us to express \emph{data policies} and \emph{application policies} and perform reasoning, in a realistic context, and also serving as the foundation for our benchmark.

To achieve this, we extended the Community Solid Server \cite{herwegen_communitysolidservercommunitysolidserver_2023} v6.0, a modular and extensible Solid server implementation written in TypeScript. For policy reasoning, we use an off-the-shelf reasoner, EYE \cite{verborgh_drawing_2015}, which supports RDF Surfaces and N3 reasoning. In particular, we used the \lstinline|eyereasoner| package available on npm, which is a WebAssembly distribution of EYE with a JavaScript interface.

\begin{figure}
    \centering
    \includegraphics[width=\linewidth]{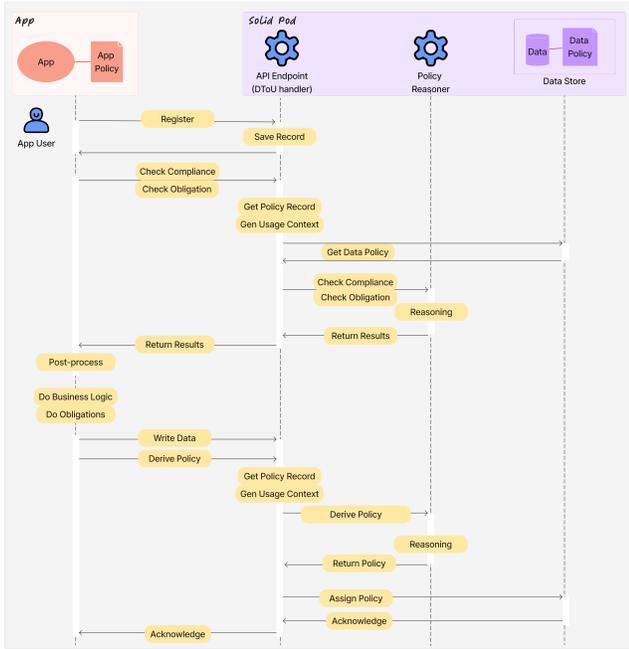}
    \caption{Sequence diagram for the Solid integration of the DToU language}
    \label{fig:seq_diagram}
\end{figure}

In the sequence diagram depicted in Figure \ref{fig:seq_diagram}, we illustrate the key components and actions related to policy reasoning for applications, assuming that the data already have DToU policies associated.

Before reasoning, the application needs to register its application policy. The DToU handler, located behind the API endpoint on the server, processes the request and creates a temporary policy record.

Subsequently, the application requests a \textbf{conformance check} before utilizing the data. The DToU handler retrieves the corresponding policy, establishes the \lstinline|UsageContext|, identifies the input data from the application policy, retrieves the relevant data policy, and calls the policy engine to perform the conformance check. The results are then provided to the application for further action, such as user display.
With a complete transformation to DToU (from current access control), the DToU handler shall deny usage of data without policies or in the presence of conflicts.

Similarly, the application can request an \textbf{obligation check} and then processes the information accordingly, including displaying it to the user at the appropriate time or automatically fulfilling obligations when applicable. Our language distinguishes between \lstinline|UserObligation| and \lstinline|ProcessObligation| classes to facilitate this distinction, leaving the specific obligations up to policy authors.

Finally, when the application intends to write output data to the Pod, it should send a \textbf{policy derivation} request along with the data. The DToU handler will then perform policy derivation and store the policy with the data. This allows subsequent applications to automatically carry out DToU reasoning when they request this data as input.
As mentioned, the DToU handler may deny usage of data without policies, thus effectively enforcing policy derivation -- if it does not perform it, the stored data will not be usable.

DToU reasoning is performed by the policy engine and DToU handler in the modified Solid service, reducing the burden on the application developers and Pod owners.
From the Pod owner's perspective, supporting DToU requires expressing relevant data policies for input data. 
For the applications, DToU support involves preparing app policies and sending policy-related requests to the (modified) Solid service while also handling responses.
These application policies can be statically attached to the project, or dynamically generated from a template, offering flexibility for different users. The optimal method for authoring app and data policies remains an open question, outside the scope of this paper.%Together with the registration step, this allows the application to have different behaviours, and thus enabling choices and negotiation with the user.
%Of course, it is an open question what is the optimal method to support authoring the app policy and data policy, and we do not discuss that in this paper.

\section{Performance Benchmark}

We conducted benchmark tests to evaluate the performance of our integration and to gain insights into its scalability across different reasoning tasks and workloads. This section presents the results and our discussion.

\subsection{Benchmark settings}

The benchmark encompasses a wide range of workloads of incorporating key variables: the number of different terms in the data policy and app policy. We varied these numbers, ranging from 10 to 1000, while keeping other variables at a fixed value of 10.
There are two exceptions that are not fixed to 10: the number of attributes, which is set to 100, and the number of inputs, which is limited to 4. We chose these values because attributes require references to them, and handling only 10 would not suffice for distribution among tags, prohibitions and obligations; the number of inputs significantly impacts performance, so we opted for a smaller yet reasonable number.
Policies are generated through a random process.

It is worth noting that the range we benchmarked should be demonstrable to what usually exists in current real-world policies. For example, \cite{breaux_eddy_2014} reviewed Facebook, Zynga and AOL policies and identified a total of 131, 190 and 75 statements. Although not equitable, each statement roughly corresponds to one term in the semantic layer and several attributes, or one term in the app policy.

For the benchmark, we utilized the WebAssembly distribution of the eyereasoner v6.9.5, which is available on NPM. We started the server using the provided file-storage configuration, without special parameters.
The benchmark tests were conducted on a consumer-level laptop, with an Intel Core i5-1135G7 (2.4 GHz) and 16GB of RAM, running on Linux kernel 6.1.55 (x86\_64). Each workload was repeated 10 times, and the time taken from sending requests to receiving results was recorded.

\subsection{Results and discussions}

\subsubsection{General results}

\begin{figure}
    \centering
    \includegraphics[width=\linewidth]{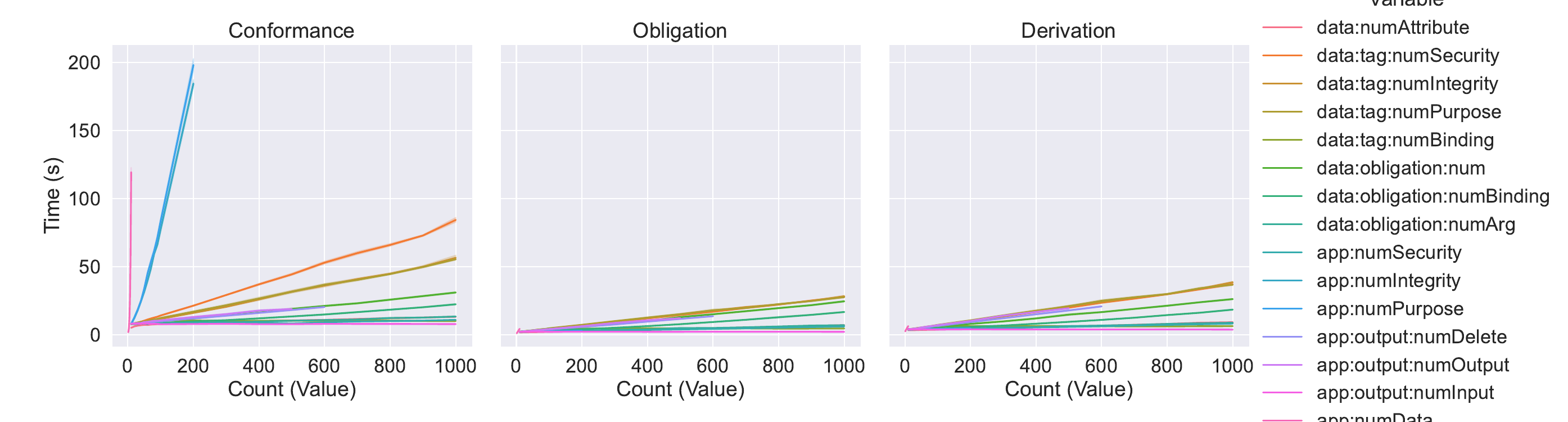}
    \caption{General benchmark results of different workloads}
    \label{fig:benchmark-0}
\end{figure}

Figure \ref{fig:benchmark-0} shows the primary results of our benchmark. Most variables exhibit a linear or sublinear scaling trend. However, some variables do not reach 1000 because they encountered time-out or connection resets with further increase.

Names of variables correspond to their specific workload. For example, \lstinline|app:output:numDelete| means the variable being changed is the number of delete refinements in each output specification of the app policy.

\begin{figure}
    \centering
    \includegraphics[width=\linewidth]{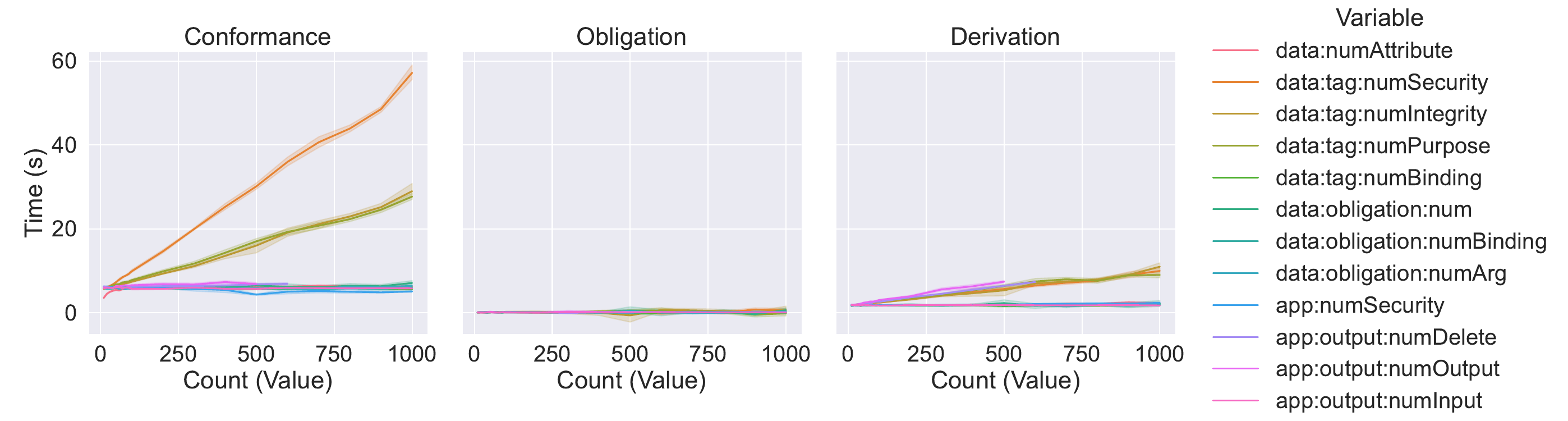}
    \caption{Selective benchmark results for most variables, subtracting the base reasoning time for general axioms}
    \label{fig:benchmark-1}
\end{figure}

The three specific variables, namely \lstinline|app:numData|, \lstinline|app:numPurpose| and \lstinline|app:numIntegrity|, displayed exceptional growing trends in conformance checking, resulting in time-outs, which we will discuss separately later. After removing these exceptional cases, and subtracting the base time for policy loading and general reasoning (e.g.~axioms for \lstinline|rdfs:subClassOf|), we obtain Figure \ref{fig:benchmark-1}. It demonstrates a clearer linear growth trend in the remaining variables.

Among them, most variables reached 1000, except for \lstinline|app:output:numDelete| and \lstinline|app:output:numOutput|, which stopped at 600 and 500. This limitation was due to connection resets, caused by the server encountering JavaScript out-of-heap errors during the 10 repetitions. This indicates a potential area for optimization of the memory consumption. However, it is worth noting that having 500 output ports or 600 delete refinements per output port (thus 6000 in total) for an application is exceptionally large and likely not an issue in realistic scenarios.

\subsubsection{Unmatched expectations}

\begin{figure}
    \centering
    \includegraphics[width=\linewidth]{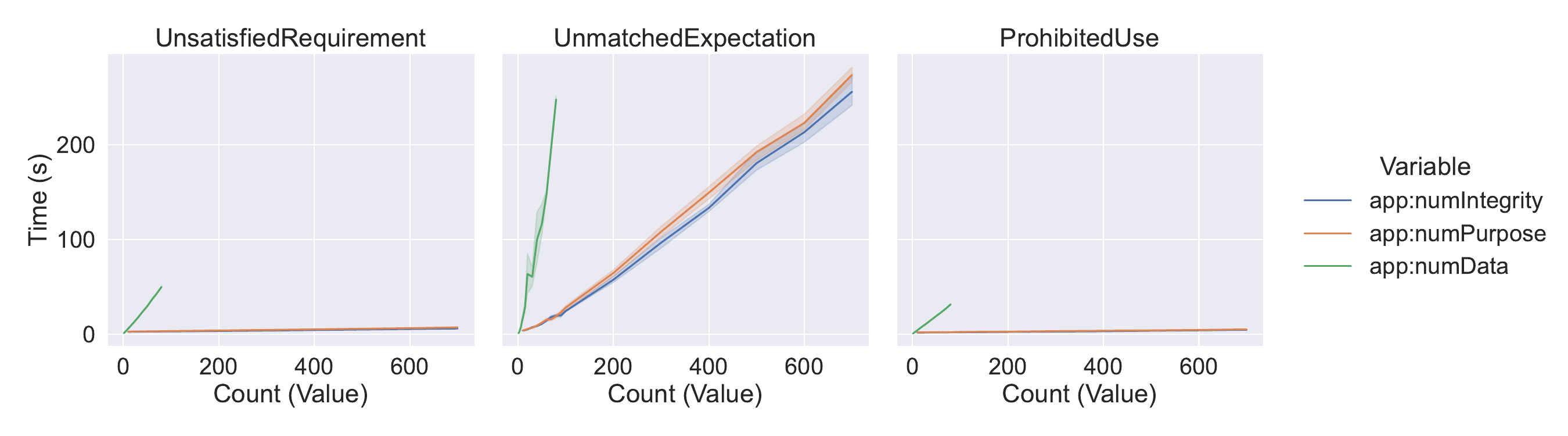}
    \caption{Experiment results for different conformance checking tasks for exceptional variables, \lstinline|app:numData|, \lstinline|app:numPurpose| and \lstinline|app:numIntegrity|}
    \label{fig:benchmark-2}
\end{figure}

Now, regarding the exceptional growths, as shown in Figure \ref{fig:benchmark-0}, two variables, \lstinline|app:numPurpose| and \lstinline|app:numIntegrity|, correspond to the checking of unmatched expectations. We further performed additional experiments to understand the time spent on different conformance checking (sub-)tasks, as shown in Figure \ref{fig:benchmark-2}. It verified the intuition that the time was mainly consumed by checking the unmatched expectations.

However, it is interesting to note that the time for its symmetric task, unsatisfied requirements, did not exhibit the same exceptional growth, as seen in Figure \ref{fig:benchmark-0} for variable \lstinline|data:tag:numSecurity| and \lstinline|app:numSecurity|. This suggests that the issue is likely not due to a mis-implementation of our axioms, but rather a performance bottleneck for RDF Surfaces or the EYE reasoner related to specific rule combinations and orders. Ideally, future work should explore and address this issue, with the potential to further optimize by reducing the coefficient for tag matching. 

%Ideally, future work can try to explore and address this issue. Having said that, it is also roughly a linear scalability trend, showing the potential to be optimized further by reducing the coefficient.

\subsubsection{Number of data / inputs}

Changing the number of data inputs led to the most significant growth in reasoning time, displaying a growth pattern that appears higher than linear, as shown in the sub-figure for unmatched expectations in Figure \ref{fig:benchmark-2}. This could be explained because all reasoning tasks depend on the pairing of data policies and inputs, making the complexity increase faster with the number of data inputs. However, it is interesting to note that obligation checking and policy derivation did not pose the same exceptional growth trend as conformance checking in Figure \ref{fig:benchmark-0}.
A plausible explanation is that conformance check involves a heavier workload, making this effect more pronounced. This is supported by the fact that the time spent on a more complex subtask, such as unmatched expectation, also grew faster than that of unsatisfied requirements and prohibited use, as reflected in Figure \ref{fig:benchmark-2}.
This suggests that our usage of RDF Surfaces may be suboptimal and should be addressed and optimized in future work.

\subsection{Conclusion}

From our benchmark, it is clear that the reasoning cost for all tasks shows a linear growth concerning all but one factor. It is important to note that DToU policy reasoning is not performed frequently (typically three possibilities in an application's lifecycle), so real-time performance is not a strict requirement. However, our result highlight the significant potential for deploying our DToU language on a large scale with a substantial volume of policies. Additionally, we delved into the specifics of the suboptimal results and proposed potential reasons for their behaviour. In a production system, it is essential to focus on optimizing the reasoning tasks and the underlying reasoner to ensure efficient and reliable performance.

%We also looked into details of the suboptimal results, and proposed potential reasons for their behaviour. In general, in a production system, optimization of the reasoning task and reasoner needs to be performed.

\section{Summary and Future Work}

In this work, we have undertaken a comprehensive exploration of the challenges and advantages associated with introducing DToU into a decentralized Web context, exemplified by Solid, with the overarching aim of enhancing user autonomy. Our efforts have included identifying the pertinent challenges and benefits, delineating the specific requirements for a policy language within this context, and introduced our own DToU policy language, which is based on semantic technologies. We have also detailed the design of data policies, application policies and the underlying reasoning mechanism. Furthermore, we showed how our solution integrates with Solid, along with benchmark tests that assessed the scalability of our implementation, highlighting its potential for wider adoption with a substantial volume of policies. We discussed areas where further optimization is required. 

The next step of our work involves evaluating the language expressiveness and its understandability for users. We are also interested in exploring simpler methods for policy authorization, which could potentially involve leveraging NLP technologies.
In general, our work underscores a paradigm shift in how application may be selected, permission granted, and interoperability achieved. This shift is driven by the automated reasoning of \emph{perennial} DToU policies. It points out a wide spectrum of challenges and opportunities, including but not limited to enhancing the language expressiveness, improving usability, effectively maintaining policies, and optimizing performance.
Effectively addressing these social-technical challenges will require interdisciplinary collaboration, and we are committed to contributing to this ongoing effort.

%%
%% The acknowledgments section is defined using the "acks" environment
%% (and NOT an unnumbered section). This ensures the proper
%% identification of the section in the article metadata, and the
%% consistent spelling of the heading.
\begin{acks}
This research was supported by the Oxford Martin School Programme for Ethical Web and Data Architectures in the Age of AI at the University of Oxford. Special thanks to Nigel Shadbolt, Tim Berners-Lee and Ruben Verborgh for their engagement and comments in early discussions of this work.
\end{acks}

%%
%% The next two lines define the bibliography style to be used, and
%% the bibliography file.
\bibliographystyle{ACM-Reference-Format}
\balance
\bibliography{main,extra}

%%% -*-BibTeX-*-
%%% Do NOT edit. File created by BibTeX with style
%%% ACM-Reference-Format-Journals [18-Jan-2012].

\begin{thebibliography}{40}

%%% ====================================================================
%%% NOTE TO THE USER: you can override these defaults by providing
%%% customized versions of any of these macros before the \bibliography
%%% command.  Each of them MUST provide its own final punctuation,
%%% except for \shownote{}, \showDOI{}, and \showURL{}.  The latter two
%%% do not use final punctuation, in order to avoid confusing it with
%%% the Web address.
%%%
%%% To suppress output of a particular field, define its macro to expand
%%% to an empty string, or better, \unskip, like this:
%%%
%%% \newcommand{\showDOI}[1]{\unskip}   % LaTeX syntax
%%%
%%% \def \showDOI #1{\unskip}           % plain TeX syntax
%%%
%%% ====================================================================

\ifx \showCODEN    \undefined \def \showCODEN     #1{\unskip}     \fi
\ifx \showDOI      \undefined \def \showDOI       #1{#1}\fi
\ifx \showISBNx    \undefined \def \showISBNx     #1{\unskip}     \fi
\ifx \showISBNxiii \undefined \def \showISBNxiii  #1{\unskip}     \fi
\ifx \showISSN     \undefined \def \showISSN      #1{\unskip}     \fi
\ifx \showLCCN     \undefined \def \showLCCN      #1{\unskip}     \fi
\ifx \shownote     \undefined \def \shownote      #1{#1}          \fi
\ifx \showarticletitle \undefined \def \showarticletitle #1{#1}   \fi
\ifx \showURL      \undefined \def \showURL       {\relax}        \fi
% The following commands are used for tagged output and should be
% invisible to TeX
\providecommand\bibfield[2]{#2}
\providecommand\bibinfo[2]{#2}
\providecommand\natexlab[1]{#1}
\providecommand\showeprint[2][]{arXiv:#2}

\bibitem[noa(2013)]%
        {noauthor_extensible_2013}
 \bibinfo{year}{2013}\natexlab{}.
\newblock \bibinfo{title}{{eXtensible} {Access} {Control} {Markup} {Language} ({XACML}) {Version} 3.0}.
\newblock
\newblock
\urldef\tempurl%
\url{https://docs.oasis-open.org/xacml/3.0/xacml-3.0-core-spec-os-en.html}
\showURL{%
\tempurl}


\bibitem[noa(2018)]%
        {noauthor_odrl_2018}
 \bibinfo{year}{2018}\natexlab{}.
\newblock \bibinfo{title}{{ODRL} {Information} {Model} 2.2}.
\newblock
\newblock
\urldef\tempurl%
\url{https://www.w3.org/TR/odrl-model/}
\showURL{%
\tempurl}


\bibitem[noa(2022a)]%
        {noauthor_access_2022}
 \bibinfo{year}{2022}\natexlab{a}.
\newblock \bibinfo{title}{Access {Control} {Policy} ({ACP})}.
\newblock
\newblock
\urldef\tempurl%
\url{https://solid.github.io/authorization-panel/acp-specification/}
\showURL{%
\tempurl}


\bibitem[noa(2022b)]%
        {noauthor_web_2022}
 \bibinfo{year}{2022}\natexlab{b}.
\newblock \bibinfo{title}{Web {Access} {Control}}.
\newblock
\newblock
\urldef\tempurl%
\url{https://solid.github.io/web-access-control-spec/}
\showURL{%
\tempurl}


\bibitem[Beckett et~al\mbox{.}(2014)]%
        {beckett_rdf_2014}
\bibfield{author}{\bibinfo{person}{David Beckett}, \bibinfo{person}{{Tim Berners-Lee}}, \bibinfo{person}{{Eric Prud'hommeaux}}, {and} \bibinfo{person}{{Gavin Carothers}}.} \bibinfo{year}{2014}\natexlab{}.
\newblock \bibinfo{title}{{RDF} 1.1 {Turtle}}.
\newblock
\newblock
\urldef\tempurl%
\url{https://www.w3.org/TR/turtle/}
\showURL{%
\tempurl}


\bibitem[Berners-Lee et~al\mbox{.}(2008)]%
        {berners-lee_n3logic_2008}
\bibfield{author}{\bibinfo{person}{Tim Berners-Lee}, \bibinfo{person}{Dan Connolly}, \bibinfo{person}{Lalana Kagal}, \bibinfo{person}{Yosi Scharf}, {and} \bibinfo{person}{Jim Hendler}.} \bibinfo{year}{2008}\natexlab{}.
\newblock \showarticletitle{{N3Logic}: {A} logical framework for the {World} {Wide} {Web}}.
\newblock \bibinfo{journal}{\emph{Theory and Practice of Logic Programming}} \bibinfo{volume}{8}, \bibinfo{number}{3} (\bibinfo{date}{May} \bibinfo{year}{2008}), \bibinfo{pages}{249--269}.
\newblock
\showISSN{1475-3081, 1471-0684}
\urldef\tempurl%
\url{https://doi.org/10.1017/S1471068407003213}
\showDOI{\tempurl}
\newblock
\shownote{Publisher: Cambridge University Press}.


\bibitem[Biswas et~al\mbox{.}(2016)]%
        {biswas_label-based_2016}
\bibfield{author}{\bibinfo{person}{Prosunjit Biswas}, \bibinfo{person}{Ravi Sandhu}, {and} \bibinfo{person}{Ram Krishnan}.} \bibinfo{year}{2016}\natexlab{}.
\newblock \showarticletitle{Label-{Based} {Access} {Control}: {An} {ABAC} {Model} with {Enumerated} {Authorization} {Policy}}. In \bibinfo{booktitle}{\emph{Proceedings of the 2016 {ACM} {International} {Workshop} on {Attribute} {Based} {Access} {Control}}} \emph{(\bibinfo{series}{{ABAC} '16})}. \bibinfo{publisher}{Association for Computing Machinery}, \bibinfo{address}{New York, NY, USA}, \bibinfo{pages}{1--12}.
\newblock
\showISBNx{978-1-4503-4079-3}
\urldef\tempurl%
\url{https://doi.org/10.1145/2875491.2875498}
\showDOI{\tempurl}


\bibitem[Breaux et~al\mbox{.}(2014)]%
        {breaux_eddy_2014}
\bibfield{author}{\bibinfo{person}{Travis~D. Breaux}, \bibinfo{person}{Hanan Hibshi}, {and} \bibinfo{person}{Ashwini Rao}.} \bibinfo{year}{2014}\natexlab{}.
\newblock \showarticletitle{Eddy, a formal language for specifying and analyzing data flow specifications for conflicting privacy requirements}.
\newblock \bibinfo{journal}{\emph{Requirements Engineering}} \bibinfo{volume}{19}, \bibinfo{number}{3} (\bibinfo{date}{Sept.} \bibinfo{year}{2014}), \bibinfo{pages}{281--307}.
\newblock
\showISSN{1432-010X}
\urldef\tempurl%
\url{https://doi.org/10.1007/s00766-013-0190-7}
\showDOI{\tempurl}


\bibitem[Elnikety et~al\mbox{.}(2016)]%
        {elnikety_thoth_2016}
\bibfield{author}{\bibinfo{person}{Eslam Elnikety}, \bibinfo{person}{Aastha Mehta}, \bibinfo{person}{Anjo Vahldiek-Oberwagner}, \bibinfo{person}{Deepak Garg}, {and} \bibinfo{person}{Peter Druschel}.} \bibinfo{year}{2016}\natexlab{}.
\newblock \showarticletitle{Thoth: {Comprehensive} {Policy} {Compliance} in {Data} {Retrieval} {Systems}}. In \bibinfo{booktitle}{\emph{Proceedings of the 25th {USENIX} {Conference} on {Security} {Symposium}}} \emph{(\bibinfo{series}{{SEC}'16})}. \bibinfo{publisher}{USENIX Association}, \bibinfo{address}{Berkeley, CA, USA}, \bibinfo{pages}{637--654}.
\newblock
\showISBNx{978-1-931971-32-4}
\urldef\tempurl%
\url{https://www.usenix.org/conference/usenixsecurity16/technical-sessions/presentation/elnikety}
\showURL{%
\tempurl}


\bibitem[Esteves et~al\mbox{.}(2021)]%
        {esteves_odrl_2021}
\bibfield{author}{\bibinfo{person}{Beatriz Esteves}, \bibinfo{person}{Harshvardhan~J. Pandit}, {and} \bibinfo{person}{Víctor Rodríguez-Doncel}.} \bibinfo{year}{2021}\natexlab{}.
\newblock \showarticletitle{{ODRL} {Profile} for {Expressing} {Consent} through {Granular} {Access} {Control} {Policies} in {Solid}}. In \bibinfo{booktitle}{\emph{2021 {IEEE} {European} {Symposium} on {Security} and {Privacy} {Workshops} ({EuroS}\&{PW})}}. \bibinfo{pages}{298--306}.
\newblock
\urldef\tempurl%
\url{https://doi.org/10.1109/EuroSPW54576.2021.00038}
\showDOI{\tempurl}
\newblock
\shownote{ISSN: 2768-0657}.


\bibitem[Herwegen et~al\mbox{.}(2023)]%
        {herwegen_communitysolidservercommunitysolidserver_2023}
\bibfield{author}{\bibinfo{person}{Joachim~Van Herwegen}, \bibinfo{person}{Ruben Verborgh}, \bibinfo{person}{Ruben Taelman}, \bibinfo{person}{Thomas Dupont}, \bibinfo{person}{Matthieu Bosquet}, \bibinfo{person}{Mend Renovate}, \bibinfo{person}{Jasper Vaneessen}, \bibinfo{person}{smessie}, \bibinfo{person}{Arthur Joppart}, \bibinfo{person}{wkerckho}, \bibinfo{person}{Simone Persiani}, \bibinfo{person}{Wouter Termont}, \bibinfo{person}{Michiel~de Jong}, \bibinfo{person}{Noel~De Martin}, \bibinfo{person}{Stijn Taelemans}, \bibinfo{person}{Wout Slabbinck}, \bibinfo{person}{Jesse Wright}, \bibinfo{person}{jaxoncreed}, \bibinfo{person}{surilindur}, \bibinfo{person}{zg009}, \bibinfo{person}{Aaron Coburn}, \bibinfo{person}{Adler Faulkner}, \bibinfo{person}{Brandon Aaron}, \bibinfo{person}{Charlie Blevins}, \bibinfo{person}{Dylan~Van Assche}, \bibinfo{person}{Emelia Smith}, \bibinfo{person}{Freya}, {and} \bibinfo{person}{Gertjan~De Mulder}.} \bibinfo{year}{2023}\natexlab{}.
\newblock \bibinfo{title}{{CommunitySolidServer}/{CommunitySolidServer}}.
\newblock
\newblock
\urldef\tempurl%
\url{https://doi.org/10.5281/zenodo.8410285}
\showDOI{\tempurl}


\bibitem[Hochstenbach et~al\mbox{.}(2023)]%
        {hochstenbach_rdf_2023}
\bibfield{author}{\bibinfo{person}{Patrick Hochstenbach}, \bibinfo{person}{Jos De~Roo}, {and} \bibinfo{person}{Ruben Verborgh}.} \bibinfo{year}{2023}\natexlab{}.
\newblock \showarticletitle{{RDF} {Surfaces}: {Computer} {Says} {No}}.
\newblock
\urldef\tempurl%
\url{http://arxiv.org/abs/2305.08476}
\showURL{%
\tempurl}
\newblock
\shownote{arXiv:2305.08476 [cs]}.


\bibitem[Huh-Yoo and Rader(2020)]%
        {huh-yoo_its_2020}
\bibfield{author}{\bibinfo{person}{Jina Huh-Yoo} {and} \bibinfo{person}{Emilee Rader}.} \bibinfo{year}{2020}\natexlab{}.
\newblock \showarticletitle{It's the {Wild}, {Wild} {West}: {Lessons} {Learned} {From} {IRB} {Members}' {Risk} {Perceptions} {Toward} {Digital} {Research} {Data}}.
\newblock \bibinfo{journal}{\emph{Proceedings of the ACM on Human-Computer Interaction}} \bibinfo{volume}{4}, \bibinfo{number}{CSCW1} (\bibinfo{date}{May} \bibinfo{year}{2020}), \bibinfo{pages}{059:1--059:22}.
\newblock
\urldef\tempurl%
\url{https://doi.org/10.1145/3392868}
\showDOI{\tempurl}


\bibitem[{Håvard D. Johansen} et~al\mbox{.}(2015)]%
        {havard_d_johansen_enforcing_2015}
\bibfield{author}{\bibinfo{person}{{Håvard D. Johansen}}, \bibinfo{person}{{Eleanor Birrell}}, \bibinfo{person}{{Robbert van Renesse}}, \bibinfo{person}{{Fred B. Schneider}}, \bibinfo{person}{{Magnus Stenhaug}}, {and} \bibinfo{person}{{Dag Johansen}}.} \bibinfo{year}{2015}\natexlab{}.
\newblock \showarticletitle{Enforcing {Privacy} {Policies} with {Meta}-{Code}}. In \bibinfo{booktitle}{\emph{Proceedings of the 6th {Asia}-{Pacific} {Workshop} on {Systems}}} \emph{(\bibinfo{series}{{APSys} '15})}. \bibinfo{publisher}{ACM Press}, \bibinfo{address}{Tokyo, Japan}, \bibinfo{pages}{1--7}.
\newblock
\showISBNx{978-1-4503-3554-6}
\urldef\tempurl%
\url{https://doi.org/10.1145/2797022.2797040}
\showDOI{\tempurl}


\bibitem[Ion et~al\mbox{.}(2011)]%
        {ion_home_2011}
\bibfield{author}{\bibinfo{person}{Iulia Ion}, \bibinfo{person}{Niharika Sachdeva}, \bibinfo{person}{Ponnurangam Kumaraguru}, {and} \bibinfo{person}{Srdjan Čapkun}.} \bibinfo{year}{2011}\natexlab{}.
\newblock \showarticletitle{Home is safer than the cloud! privacy concerns for consumer cloud storage}. In \bibinfo{booktitle}{\emph{Proceedings of the {Seventh} {Symposium} on {Usable} {Privacy} and {Security}}} \emph{(\bibinfo{series}{{SOUPS} '11})}. \bibinfo{publisher}{Association for Computing Machinery}, \bibinfo{address}{New York, NY, USA}, \bibinfo{pages}{1--20}.
\newblock
\showISBNx{978-1-4503-0911-0}
\urldef\tempurl%
\url{https://doi.org/10.1145/2078827.2078845}
\showDOI{\tempurl}


\bibitem[Iyilade and Vassileva(2014)]%
        {iyilade_p2u_2014}
\bibfield{author}{\bibinfo{person}{Johnson Iyilade} {and} \bibinfo{person}{Julita Vassileva}.} \bibinfo{year}{2014}\natexlab{}.
\newblock \showarticletitle{{P2U}: {A} {Privacy} {Policy} {Specification} {Language} for {Secondary} {Data} {Sharing} and {Usage}}. In \bibinfo{booktitle}{\emph{2014 {IEEE} {Security} and {Privacy} {Workshops}}}. \bibinfo{pages}{18--22}.
\newblock
\urldef\tempurl%
\url{https://doi.org/10.1109/SPW.2014.12}
\showDOI{\tempurl}


\bibitem[Kagal et~al\mbox{.}(2008)]%
        {kagal_using_2008}
\bibfield{author}{\bibinfo{person}{Lalana Kagal}, \bibinfo{person}{Chris Hanson}, {and} \bibinfo{person}{Daniel Weitzner}.} \bibinfo{year}{2008}\natexlab{}.
\newblock \showarticletitle{Using {Dependency} {Tracking} to {Provide} {Explanations} for {Policy} {Management}}. In \bibinfo{booktitle}{\emph{2008 {IEEE} {Workshop} on {Policies} for {Distributed} {Systems} and {Networks}}}. \bibinfo{pages}{54--61}.
\newblock
\urldef\tempurl%
\url{https://doi.org/10.1109/POLICY.2008.51}
\showDOI{\tempurl}


\bibitem[Karjoth et~al\mbox{.}(2002)]%
        {karjoth_platform_2002}
\bibfield{author}{\bibinfo{person}{Günter Karjoth}, \bibinfo{person}{Matthias Schunter}, {and} \bibinfo{person}{Michael Waidner}.} \bibinfo{year}{2002}\natexlab{}.
\newblock \showarticletitle{Platform for {Enterprise} {Privacy} {Practices}: {Privacy}-{Enabled} {Management} of {Customer} {Data}}. In \bibinfo{booktitle}{\emph{Privacy {Enhancing} {Technologies}}} \emph{(\bibinfo{series}{Lecture {Notes} in {Computer} {Science}})}. \bibinfo{publisher}{Springer, Berlin, Heidelberg}, \bibinfo{pages}{69--84}.
\newblock
\showISBNx{978-3-540-00565-0}
\urldef\tempurl%
\url{https://doi.org/10.1007/3-540-36467-6_6}
\showDOI{\tempurl}


\bibitem[Lazouski et~al\mbox{.}(2010)]%
        {lazouski_usage_2010}
\bibfield{author}{\bibinfo{person}{Aliaksandr Lazouski}, \bibinfo{person}{Fabio Martinelli}, {and} \bibinfo{person}{Paolo Mori}.} \bibinfo{year}{2010}\natexlab{}.
\newblock \showarticletitle{Usage control in computer security: {A} survey}.
\newblock \bibinfo{journal}{\emph{Computer Science Review}} \bibinfo{volume}{4}, \bibinfo{number}{2} (\bibinfo{date}{May} \bibinfo{year}{2010}), \bibinfo{pages}{81--99}.
\newblock
\showISSN{1574-0137}
\urldef\tempurl%
\url{https://doi.org/10.1016/j.cosrev.2010.02.002}
\showDOI{\tempurl}


\bibitem[McDonald and Cranor(2008)]%
        {mcdonald_cost_2008}
\bibfield{author}{\bibinfo{person}{Aleecia~M McDonald} {and} \bibinfo{person}{Lorrie~Faith Cranor}.} \bibinfo{year}{2008}\natexlab{}.
\newblock \showarticletitle{The cost of reading privacy policies}.
\newblock \bibinfo{journal}{\emph{Isjlp}}  \bibinfo{volume}{4} (\bibinfo{year}{2008}), \bibinfo{pages}{543}.
\newblock
\newblock
\shownote{Publisher: HeinOnline}.


\bibitem[Meurisch et~al\mbox{.}(2020)]%
        {meurisch_privacy-preserving_2020}
\bibfield{author}{\bibinfo{person}{Christian Meurisch}, \bibinfo{person}{Bekir Bayrak}, {and} \bibinfo{person}{Max Mühlhäuser}.} \bibinfo{year}{2020}\natexlab{}.
\newblock \showarticletitle{Privacy-preserving {AI} {Services} {Through} {Data} {Decentralization}}. In \bibinfo{booktitle}{\emph{Proceedings of {The} {Web} {Conference} 2020}}. \bibinfo{publisher}{Association for Computing Machinery}, \bibinfo{address}{New York, NY, USA}, \bibinfo{pages}{190--200}.
\newblock
\showISBNx{978-1-4503-7023-3}
\urldef\tempurl%
\url{http://doi.org/10.1145/3366423.3380106}
\showURL{%
\tempurl}


\bibitem[Mont et~al\mbox{.}(2003)]%
        {mont_towards_2003}
\bibfield{author}{\bibinfo{person}{M.~C. Mont}, \bibinfo{person}{S. Pearson}, {and} \bibinfo{person}{P. Bramhall}.} \bibinfo{year}{2003}\natexlab{}.
\newblock \showarticletitle{Towards accountable management of identity and privacy: sticky policies and enforceable tracing services}. In \bibinfo{booktitle}{\emph{14th {International} {Workshop} on {Database} and {Expert} {Systems} {Applications}, 2003. {Proceedings}.}} \bibinfo{pages}{377--382}.
\newblock
\urldef\tempurl%
\url{https://doi.org/10.1109/DEXA.2003.1232051}
\showDOI{\tempurl}


\bibitem[Myers and Liskov(1997)]%
        {myers_decentralized_1997}
\bibfield{author}{\bibinfo{person}{Andrew~C. Myers} {and} \bibinfo{person}{Barbara Liskov}.} \bibinfo{year}{1997}\natexlab{}.
\newblock \showarticletitle{A {Decentralized} {Model} for {Information} {Flow} {Control}}. In \bibinfo{booktitle}{\emph{Proceedings of the {Sixteenth} {ACM} {Symposium} on {Operating} {Systems} {Principles}}} \emph{(\bibinfo{series}{{SOSP} '97})}. \bibinfo{publisher}{ACM}, \bibinfo{address}{New York, NY, USA}, \bibinfo{pages}{129--142}.
\newblock
\showISBNx{978-0-89791-916-6}
\urldef\tempurl%
\url{https://doi.org/10.1145/268998.266669}
\showDOI{\tempurl}


\bibitem[Obar and Oeldorf-Hirsch(2020)]%
        {obar_biggest_2020}
\bibfield{author}{\bibinfo{person}{Jonathan~A. Obar} {and} \bibinfo{person}{Anne Oeldorf-Hirsch}.} \bibinfo{year}{2020}\natexlab{}.
\newblock \showarticletitle{The biggest lie on the {Internet}: ignoring the privacy policies and terms of service policies of social networking services}.
\newblock \bibinfo{journal}{\emph{Information, Communication \& Society}} \bibinfo{volume}{23}, \bibinfo{number}{1} (\bibinfo{date}{Jan.} \bibinfo{year}{2020}), \bibinfo{pages}{128--147}.
\newblock
\showISSN{1369-118X}
\urldef\tempurl%
\url{https://doi.org/10.1080/1369118X.2018.1486870}
\showDOI{\tempurl}


\bibitem[Pasquier et~al\mbox{.}(2017)]%
        {pasquier_camflow_2017}
\bibfield{author}{\bibinfo{person}{Thomas F. J.-M. Pasquier}, \bibinfo{person}{Jatinder Singh}, \bibinfo{person}{David Eyers}, {and} \bibinfo{person}{Jean Bacon}.} \bibinfo{year}{2017}\natexlab{}.
\newblock \showarticletitle{{CamFlow}: {Managed} {Data}-sharing for {Cloud} {Services}}.
\newblock \bibinfo{journal}{\emph{IEEE Transactions on Cloud Computing}} \bibinfo{volume}{5}, \bibinfo{number}{3} (\bibinfo{date}{July} \bibinfo{year}{2017}), \bibinfo{pages}{472--484}.
\newblock
\showISSN{2168-7161}
\urldef\tempurl%
\url{https://doi.org/10.1109/TCC.2015.2489211}
\showDOI{\tempurl}
\newblock
\shownote{arXiv: 1506.04391}.


\bibitem[Pearson and Casassa-Mont(2011)]%
        {pearson_sticky_2011}
\bibfield{author}{\bibinfo{person}{S. Pearson} {and} \bibinfo{person}{M. Casassa-Mont}.} \bibinfo{year}{2011}\natexlab{}.
\newblock \showarticletitle{Sticky {Policies}: {An} {Approach} for {Managing} {Privacy} across {Multiple} {Parties}}.
\newblock \bibinfo{journal}{\emph{Computer}} \bibinfo{volume}{44}, \bibinfo{number}{9} (\bibinfo{date}{Sept.} \bibinfo{year}{2011}), \bibinfo{pages}{60--68}.
\newblock
\showISSN{0018-9162}
\urldef\tempurl%
\url{https://doi.org/10.1109/MC.2011.225}
\showDOI{\tempurl}


\bibitem[Qiu et~al\mbox{.}(2020)]%
        {qiu_survey_2020}
\bibfield{author}{\bibinfo{person}{Jing Qiu}, \bibinfo{person}{Zhihong Tian}, \bibinfo{person}{Chunlai Du}, \bibinfo{person}{Qi Zuo}, \bibinfo{person}{Shen Su}, {and} \bibinfo{person}{Binxing Fang}.} \bibinfo{year}{2020}\natexlab{}.
\newblock \showarticletitle{A {Survey} on {Access} {Control} in the {Age} of {Internet} of {Things}}.
\newblock \bibinfo{journal}{\emph{IEEE Internet of Things Journal}} \bibinfo{volume}{7}, \bibinfo{number}{6} (\bibinfo{date}{June} \bibinfo{year}{2020}), \bibinfo{pages}{4682--4696}.
\newblock
\showISSN{2327-4662}
\urldef\tempurl%
\url{https://doi.org/10.1109/JIOT.2020.2969326}
\showDOI{\tempurl}
\newblock
\shownote{Conference Name: IEEE Internet of Things Journal}.


\bibitem[{Rui Zhao} et~al\mbox{.}(2021)]%
        {rui_zhao_draid_2021}
\bibfield{author}{\bibinfo{person}{{Rui Zhao}}, \bibinfo{person}{{Malcolm Atkinson}}, \bibinfo{person}{{Petros Papapanagiotou}}, \bibinfo{person}{{Federica Magnoni}}, {and} \bibinfo{person}{{Jacques Fleuriot}}.} \bibinfo{year}{2021}\natexlab{}.
\newblock \showarticletitle{Dr.{Aid}: {Supporting} {Data}-governance {Rule} {Compliance} for {Decentralized} {Collaboration} in an {Automated} {Way}}. In \bibinfo{booktitle}{\emph{The 24th {ACM} {Conference} on {Computer}-{Supported} {Cooperative} {Work} and {Social} {Computing} ({CSCW})}}.
\newblock
\urldef\tempurl%
\url{https://doi.org/10.1145/3479604}
\showDOI{\tempurl}


\bibitem[Sagirlar et~al\mbox{.}(2018)]%
        {sagirlar_decentralizing_2018}
\bibfield{author}{\bibinfo{person}{Gokhan Sagirlar}, \bibinfo{person}{Barbara Carminati}, {and} \bibinfo{person}{Elena Ferrari}.} \bibinfo{year}{2018}\natexlab{}.
\newblock \showarticletitle{Decentralizing privacy enforcement for {Internet} of {Things} smart objects}.
\newblock \bibinfo{journal}{\emph{Computer Networks}}  \bibinfo{volume}{143} (\bibinfo{date}{Oct.} \bibinfo{year}{2018}), \bibinfo{pages}{112--125}.
\newblock
\showISSN{1389-1286}
\urldef\tempurl%
\url{https://doi.org/10.1016/j.comnet.2018.07.019}
\showDOI{\tempurl}


\bibitem[Sambra et~al\mbox{.}(2016)]%
        {sambra_solid_2016}
\bibfield{author}{\bibinfo{person}{A. Sambra}, \bibinfo{person}{Essam Mansour}, \bibinfo{person}{Sandro Hawke}, \bibinfo{person}{Maged Zereba}, \bibinfo{person}{Nicola Greco}, \bibinfo{person}{Abdurrahman Ghanem}, \bibinfo{person}{D. Zagidulin}, \bibinfo{person}{Ashraf Aboulnaga}, {and} \bibinfo{person}{T. Berners-Lee}.} \bibinfo{year}{2016}\natexlab{}.
\newblock \showarticletitle{Solid: {A} {Platform} for {Decentralized} {Social} {Applications} {Based} on {Linked} {Data}}.
\newblock \bibinfo{journal}{\emph{MIT CSAIL \& Qatar Computing Research Institute, Tech. Rep.}} (\bibinfo{year}{2016}).
\newblock
\urldef\tempurl%
\url{https://www.semanticscholar.org/paper/Solid-%3A-A-Platform-for-Decentralized-Social-Based-Sambra-Mansour/5ac93548fd0628f7ff8ff65b5878d04c79c513c4}
\showURL{%
\tempurl}


\bibitem[Sandhu et~al\mbox{.}(1996)]%
        {sandhu_role-based_1996}
\bibfield{author}{\bibinfo{person}{R.S. Sandhu}, \bibinfo{person}{E.J. Coyne}, \bibinfo{person}{H.L. Feinstein}, {and} \bibinfo{person}{C.E. Youman}.} \bibinfo{year}{1996}\natexlab{}.
\newblock \showarticletitle{Role-based access control models}.
\newblock \bibinfo{journal}{\emph{Computer}} \bibinfo{volume}{29}, \bibinfo{number}{2} (\bibinfo{date}{Feb.} \bibinfo{year}{1996}), \bibinfo{pages}{38--47}.
\newblock
\showISSN{1558-0814}
\urldef\tempurl%
\url{https://doi.org/10.1109/2.485845}
\showDOI{\tempurl}
\newblock
\shownote{Conference Name: Computer}.


\bibitem[Sandhu and Park(2003)]%
        {sandhu_usage_2003}
\bibfield{author}{\bibinfo{person}{Ravi Sandhu} {and} \bibinfo{person}{Jaehong Park}.} \bibinfo{year}{2003}\natexlab{}.
\newblock \showarticletitle{Usage {Control}: {A} {Vision} for {Next} {Generation} {Access} {Control}}. In \bibinfo{booktitle}{\emph{Computer {Network} {Security}}} \emph{(\bibinfo{series}{Lecture {Notes} in {Computer} {Science}})}, \bibfield{editor}{\bibinfo{person}{Vladimir Gorodetsky}, \bibinfo{person}{Leonard Popyack}, {and} \bibinfo{person}{Victor Skormin}} (Eds.). \bibinfo{publisher}{Springer}, \bibinfo{address}{Berlin, Heidelberg}, \bibinfo{pages}{17--31}.
\newblock
\showISBNx{978-3-540-45215-7}
\urldef\tempurl%
\url{https://doi.org/10.1007/978-3-540-45215-7_2}
\showDOI{\tempurl}


\bibitem[Sandhu and Samarati(1994)]%
        {sandhu_access_1994}
\bibfield{author}{\bibinfo{person}{R.S. Sandhu} {and} \bibinfo{person}{P. Samarati}.} \bibinfo{year}{1994}\natexlab{}.
\newblock \showarticletitle{Access control: principle and practice}.
\newblock \bibinfo{journal}{\emph{IEEE Communications Magazine}} \bibinfo{volume}{32}, \bibinfo{number}{9} (\bibinfo{date}{Sept.} \bibinfo{year}{1994}), \bibinfo{pages}{40--48}.
\newblock
\showISSN{1558-1896}
\urldef\tempurl%
\url{https://doi.org/10.1109/35.312842}
\showDOI{\tempurl}
\newblock
\shownote{Conference Name: IEEE Communications Magazine}.


\bibitem[Stein et~al\mbox{.}(2023)]%
        {stein_you_2023}
\bibfield{author}{\bibinfo{person}{Jake M~L Stein}, \bibinfo{person}{Vidminas Vizgirda}, \bibinfo{person}{Max Van~Kleek}, \bibinfo{person}{Reuben Binns}, \bibinfo{person}{Jun Zhao}, \bibinfo{person}{Rui Zhao}, \bibinfo{person}{Naman Goel}, \bibinfo{person}{George Chalhoub}, \bibinfo{person}{Wael~S Albayaydh}, {and} \bibinfo{person}{Nigel Shadbolt}.} \bibinfo{year}{2023}\natexlab{}.
\newblock \showarticletitle{‘{You} are you and the app. {There}’s nobody else.’: {Building} {Worker}-{Designed} {Data} {Institutions} within {Platform} {Hegemony}}. In \bibinfo{booktitle}{\emph{Proceedings of the 2023 {CHI} {Conference} on {Human} {Factors} in {Computing} {Systems}}} \emph{(\bibinfo{series}{{CHI} '23})}. \bibinfo{publisher}{Association for Computing Machinery}, \bibinfo{address}{New York, NY, USA}, \bibinfo{pages}{1--26}.
\newblock
\showISBNx{978-1-4503-9421-5}
\urldef\tempurl%
\url{https://doi.org/10.1145/3544548.3581114}
\showDOI{\tempurl}


\bibitem[Tesfay et~al\mbox{.}(2018)]%
        {tesfay_i_2018}
\bibfield{author}{\bibinfo{person}{Welderufael~B. Tesfay}, \bibinfo{person}{Peter Hofmann}, \bibinfo{person}{Toru Nakamura}, \bibinfo{person}{Shinsaku Kiyomoto}, {and} \bibinfo{person}{Jetzabel Serna}.} \bibinfo{year}{2018}\natexlab{}.
\newblock \showarticletitle{I {Read} but {Don}'t {Agree}: {Privacy} {Policy} {Benchmarking} using {Machine} {Learning} and the {EU} {GDPR}}. In \bibinfo{booktitle}{\emph{Companion {Proceedings} of the {The} {Web} {Conference} 2018}} \emph{(\bibinfo{series}{{WWW} '18})}. \bibinfo{publisher}{International World Wide Web Conferences Steering Committee}, \bibinfo{address}{Republic and Canton of Geneva, CHE}, \bibinfo{pages}{163--166}.
\newblock
\showISBNx{978-1-4503-5640-4}
\urldef\tempurl%
\url{https://doi.org/10.1145/3184558.3186969}
\showDOI{\tempurl}


\bibitem[Verborgh and De~Roo(2015)]%
        {verborgh_drawing_2015}
\bibfield{author}{\bibinfo{person}{Ruben Verborgh} {and} \bibinfo{person}{Jos De~Roo}.} \bibinfo{year}{2015}\natexlab{}.
\newblock \showarticletitle{Drawing {Conclusions} from {Linked} {Data} on the {Web}: {The} {EYE} {Reasoner}}.
\newblock \bibinfo{journal}{\emph{IEEE Software}} \bibinfo{volume}{32}, \bibinfo{number}{3} (\bibinfo{date}{May} \bibinfo{year}{2015}), \bibinfo{pages}{23--27}.
\newblock
\showISSN{1937-4194}
\urldef\tempurl%
\url{https://doi.org/10.1109/MS.2015.63}
\showDOI{\tempurl}
\newblock
\shownote{Conference Name: IEEE Software}.


\bibitem[{W3C}(2014)]%
        {w3c_rdf_2014}
\bibfield{author}{\bibinfo{person}{{W3C}}.} \bibinfo{year}{2014}\natexlab{}.
\newblock \bibinfo{title}{{RDF} 1.1 {Concepts} and {Abstract} {Syntax}}.
\newblock
\newblock
\urldef\tempurl%
\url{https://www.w3.org/TR/rdf11-concepts/}
\showURL{%
\tempurl}


\bibitem[{W3C OWL Working Group}(2012)]%
        {w3c_owl_working_group_owl_2012}
\bibfield{author}{\bibinfo{person}{{W3C OWL Working Group}}.} \bibinfo{year}{2012}\natexlab{}.
\newblock \bibinfo{title}{{OWL} 2 {Web} {Ontology} {Language} {Document} {Overview} ({Second} {Edition})}.
\newblock
\newblock
\urldef\tempurl%
\url{https://www.w3.org/TR/owl2-overview/}
\showURL{%
\tempurl}


\bibitem[Zimmeck et~al\mbox{.}(2016)]%
        {zimmeck_automated_2016}
\bibfield{author}{\bibinfo{person}{Sebastian Zimmeck}, \bibinfo{person}{Ziqi Wang}, \bibinfo{person}{Lieyong Zou}, \bibinfo{person}{Roger Iyengar}, \bibinfo{person}{Bin Liu}, \bibinfo{person}{Florian Schaub}, \bibinfo{person}{Shomir Wilson}, \bibinfo{person}{Norman Sadeh}, \bibinfo{person}{Steven Bellovin}, {and} \bibinfo{person}{Joel Reidenberg}.} \bibinfo{year}{2016}\natexlab{}.
\newblock \showarticletitle{Automated {Analysis} of {Privacy} {Requirements} for {Mobile} {Apps}}. In \bibinfo{booktitle}{\emph{2016 {AAAI} {Fall} {Symposium} {Series}}}.
\newblock
\urldef\tempurl%
\url{https://www.aaai.org/ocs/index.php/FSS/FSS16/paper/view/14113}
\showURL{%
\tempurl}


\bibitem[Zuboff(2019)]%
        {zuboff2019age}
\bibfield{author}{\bibinfo{person}{Shoshana Zuboff}.} \bibinfo{year}{2019}\natexlab{}.
\newblock \bibinfo{booktitle}{\emph{The age of surveillance capitalism: The fight for a human future at the new frontier of power: Barack Obama's books of 2019}}.
\newblock \bibinfo{publisher}{Profile books}.
\newblock


\end{thebibliography}

%%
%% If your work has an appendix, this is the place to put it.
\appendix

\section{Further on related-work table}
\label{sec:appendix:related-work-extension}

This section provides further explanations of the terms used in Table \ref{tab:related_research_features}.

For column C1, authorization (A) means a reasoning outcome that can decide whether the data usage is permitted (authorized) or not (prohibited); obligation (O) means the reasoning outcome can contain obligations (pending actions) that needs to be performed due to the specified data usage; + means the language supports more types of features (that are not usually covered in other languages), such as \emph{remedy} in ODRL.

For column C3 and C4, \checked~means the challenge is addressed by the design or feature of the language; \CheckedBox~means the challenge is addressed, but only if assuming a shared custom schema (that is not or cannot be a part of the language specification) for describing environmental information across all policies and the environment.

For columns C2, C5 and Condition, many terms should be intuitive. Specifically, capacity (C) means the properties that the data and/or the application/process need to possess to allow the data usage (e.g.~security levels); E and \textit{\textbf{E}} differs because E only refers to the \emph{finite} (types of) environmental information covered by the language specification or its foreseeable extension, while \textit{\textbf{E}} allows arbitrary environmental information; use mode (M) denotes the potential data usage modes (aka.~codes or action verbs in \cite{breaux_eddy_2014}); multi-input (I) and multi-output (O) means the policy language possesses the mechanism to handle policies from multiple inputs and multiple outputs respectively; transformation (T) means the output policy is subject to a transformation based on the input policy and the actual processing.

\section{Policy language examples}

This section presents the full form of the example policy snippets used in the paper.

\subsection{For payment info}
\label{sec:appendix:policy:data:payment-info}

Alice requires any application to respect \lstinline|banking| security level, and permit \lstinline|make-payment| and \lstinline|verify-ownership| purposes (for payment info).

\begin{lstlisting}
:attr-tag2 a :Attribute;
    :name :tag-2;
    :class :banking;
    :value :nil.
:attr-tag3 a :Attribute;
    :name :tag-3;
    :class :make-payment;
    :value :nil.
:attr-tag4 a :Attribute;
    :name :tag-4;
    :class :verify-ownership;
    :value :nil.

:attr2 a :Attribute;
    :name :det;
    :class :data-content;
    :value :payment-details.

:tag2 a :SecurityTag;
    :attribute_ref :attr-tag2;
    :validity_binding :attr2.
:tag3 a :PurposeTag;
    :attribute_ref :attr-tag3;
    :validity_binding :attr2.
:tag4 a :PurposeTag;
    :attribute_ref :attr-tag4;
    :validity_binding :attr2.
\end{lstlisting}

\subsection{For shoe size}

This policy set specifies the policy for Alice's shoe size data:

\begin{lstlisting}
:shoe-size a :Data;
    :uri <http://a.b/shoe-size>;
    :policy :policy-2.

:policy-2 a :Policy;
    :attribute :attr1;
    :obligation :ob1.   
\end{lstlisting}

\subsection{App policy set}
\label{sec:appendix:policy:app:policy-set}

\begin{lstlisting}
:app-policy a :AppPolicy;
	:name <http://happy.shop>;
	:input_spec :input1, :input2;
	:output_spec :out1.

:input2 a :InputSpec;
    :data <http://a.b/address>;
    :port [ ;name "address-in" ];
    :integrity :full-address. 
\end{lstlisting}

\subsection{Sample usage context}
\label{sec:appendix:policy:usage-context}

An example usage context for Alice to use HappyShop may be like this:

\begin{lstlisting}
:usageContext1 a :UsageContext;
	:user <http://a.b/alice#card>;
	:app [a :AppInfo; :policy :app-policy];
	:time "20230823".

:rui a :User.
\end{lstlisting}

\section{Axioms for reasoning}
\label{sec:appendix:axiom}

The axioms for reasoning are encoded as RDF Surfaces in our reasoner. For simplicity, we present the equivalent first-order logic axioms for performing reasoning for our language here. Because our policy language is based on Turtle thus RDF \cite{w3c_rdf_2014}, knowledge is represented as triples in the ABox. We employ some conventions here for the expression in first-order logic:
\begin{enumerate}
    \item $A(x, SomeType)$ denotes that the (RDF) type of entity $x$ is $SomeType$, i.e.~the triple $(x, \texttt{rdf:type}, SomeType)$ exists in the ABox;
    \item $hasPredicate(x, y)$ denotes that there is a relation $predicate$ between entity $x$ and $y$, i.e.~the triple $(x, predicate, y)$ exists in the ABox;
    \item Other predicates starting with a capital letter refer to a shorthand, which will be explained individually;
    \item Variables are universally quantified over the scope of the entire formulae, unless explicitly quantified or stated otherwise;
    \item For clarify, we write the conclusion in the beginning, similar to a Horn clause;
    \item Existential quantified entities in the conclusions denote a new node in the ABox, which is handled nicely by RDF Surfaces;
    \item Negations denote scoped negation-as-failure, implemented using N3's built-ins;
    \item We use double-line arrows to visually distinguish between the main premise and conclusions; however, they still represent material implications, same as the single-line arrows, and are encoded in RDF Surfaces as such;
    \item Namespaces are dropped from the formulae for clarity.
\end{enumerate}

\subsection{Helper axioms}

To simplify the axioms for actual reasoning tasks, we extract two common parts as helper axioms: $RelatedDataAppInput$ and \\
$InputPolicyForOutput$.

The $RelatedDataAppInput(usage, data, app, input)$ identifies and groups data policy ($data$) and corresponding input specification ($input$) together, as well as the usage context $usage$ and application policy $app$. Typically, in a reasoning, there is only one usage context ($usage$) and one application ($app$). Conformance checking often uses this predicate. It is defined as:
\begin{align*}
&RelatedDataAppInput(usage, data, app, input) \Leftarrow \\
 &\ A(usage, UsageContext) \\
 &\ \land hasApp(usage, app_s) \land hasPolicy(app_s, app) \\
 &\ \land hasInputSpec(app, input) \land A(data, Data) \land hasUri(data, uri) \\
 &\ \land hasData(input, uri)
\end{align*}
The $InputPolicyForOutput(input, policy, output)$ identifies which input (specification) and its corresponding data policy is related to an output (specification). One input has only one data policy, while one output may have multiple inputs, leading to several $InputPolicyForOutput$s. It is defined as:
\begin{align*}
&InputPolicyForOutput(input, policy, output) \Leftarrow \\
 &\ RelatedDataAppInput(usage, data, app, input) \land \\
 &\ hasOutputSpec(app, output) \land hasFrom(output, inputPort) \land \\
 &\ hasPort(input, inputPort) \land hasPolicy(data, policy)
\end{align*}

\subsection{Conformance check}
\label{sec:appendix:axiom:conformance-check}

We have explained them in the main text of the document. Therefore, this part only presents the formulae, and their explanations where necessary.

\subsubsection{Unsatisfied requirement}
% It is defined as:
%
\begin{align*}
Un&satisfiedRequirement(t, n, input) \Leftarrow \\
 &RelatedDataAppInput(usage, data, app, input) \land \\
 &hasPolicy(data, pol) \land hasRequirement(pol, req) \land \\
 &hasCategory(req, t) \land hasDescriptor(req, n) \land \\
 &\ \begin{aligned}
     \lnot \exists prov. (&hasProvide(input, prov) \land \\
     &hasCategory(prov, t) \land hasDescriptor(prov, n))
 \end{aligned}  
\end{align*}
The definition of $UnsatisfiedRequirement(t, n, input)$ is:
\begin{align*}
Un&satisfiedRequirement(t, n, input) \equiv \\
 &\exists x. A(x, UnsatisfiedRequirement) \land hasCategory(x, t) \land \\
 &hasDescriptor(x, n) \land hasInput(x, input)
\end{align*}

\subsubsection{Unmatched expectation}
% It is defined as:
%
% UnmatchedExpectation ( ?T, ?N ) <==
%     related_data_app_input( ?usage, ?data, ?app, ?input ) /\
%     {?input.expect.{type ?T; name ?N}} -->
%     ~{?data.policy.tag.{type ?T; name ?N}}
\begin{align*}
Un&matchedExpectation(t, n, input) \Leftarrow \\
 &RelatedDataAppInput(usage, data, app, input) \\
 &hasPolicy(data, pol) \land hasExpect(input, exp) \land \\
 &hasCategory(exp, t) \land hasDescriptor(exp, n) \land \\
 &\lnot \exists tag. (hasTagging(pol, tag) \land hasCategory(tag, t) \land \\
 &\quad hasDescriptor(tag, n))
\end{align*}
The definition of $UnmatchedExpectation(t, n, input)$ is:
\begin{align*}
Un&matchedExpectation(t, n, input) \equiv \\
 &\exists x. A(x, UnmatchedExpectation) \land hasCategory(x, t) \land \\
 &hasDescriptor(x, n) \land hasInput(x, input)
\end{align*}

\subsubsection{Prohibited use}
% It is defined as:
%
% ProhibitedUse ( ?M, ?N, ?P ) <==
%     related_data_app_input( ?usage, ?data, ?app, ?input ) /\
%     (
%     {?data.policy.prohibition.{mode Use; activation_condition.{app ?N; purpose ?P}}} /\ 
%     ({?app.name ?N; ?input.purpose ?P} \/ {?input.downstream.{app_name ?N; purpose ?P}})
%     )
\begin{align*}
Pr&ohibitedUse(u, n, p, input) \Leftarrow \\
 &RelatedDataAppInput(usage, data, app, input) \land \\
 &hasPolicy(data, pol) \land hasProhibition(pol, pro) \land \\
 &hasMode(pro, Use) \land hasActivationCondition(pro, ac) \land \\
 &hasUser(ac, u) \land hasApp(ac, n) \land hasPurpose(ac, p) \land ( \\
 &\ (hasUser(usage, u) \land hasName(app, n) \land hasPurpose(input, p)) \\&\ \lor \\
 &\ (hasDownstream(input, ds) \land hasAppName(ds, n) \land \\
 &\ \ hasPurpose(ds, p)) \\
 &)
\end{align*}
The definition of $ProhibitedUse(u, n, p, input)$ is:
\begin{align*}
Pr&ohibitedUse(m, n, p, input) \equiv \\
 &\exists x. A(x, ProhibitedUse) \land hasMode(x, Use) \land hasUser(x, u) \\
 &\land hasName(x, n) \land hasPurpose(x, p) \land hasInput(x, input)
\end{align*}

\subsection{Obligation check}
\label{sec:appendix:axiom:obligation-check}
% It is defined as:
% %
% % ActivatedObligation( ?ob, ?args ) <==
% %     related_data_app_input( ?usage, ?data, ?app, ?input ) /\
% %     {?data.policy.obligation.{obligation_class ?ob; args ?args; activation_condition.{user ?U; app ?N; purpose ?P}}} /\ {?usage.user ?U; ?app.name ?N; ?input.purpose ?P}
\begin{align*}
Ac&tivatedObligation( ob, args, input ) \Leftarrow \\
 &RelatedDataAppInput(usage, data, app, input) \land \\
 &hasPolicy(data, pol) \land hasObligation(pol, obl) \land \\
 &hasObligationClass(obl, ob) \land hasArgument(obl, args) \land \\
 &hasActivationCondition(obl, ac) \land hasUser(ac, u) \land \\
 &hasApp(ac, n) \land hasPurpose(ac, p) \land \\
 &hasUser(usage, u) \land hasName(app, n) \land hasPurpose(input, p)
\end{align*}
The definition of $ActivatedObligation( ob, args, input )$ is:
\begin{align*}
Ac&tivatedObligation( ob, args, input ) \equiv \\
 &\exists x. A(x, ActivatedObligation) \land hasObligationClass(x, ob) \land \\
 &hasArgument(x, args) \land hasInput(x, input)
\end{align*}
Slightly different from the conflicts, when querying activated obligations, the corresponding attributes should be returned as well.

\subsection{Policy derivation}
\label{sec:appendix:axiom:policy-derivation}

\subsubsection{Output attribute}
% Thus, we have the following axioms for handling them:
%
% OutputAttribute( ?N, ?T, ?V, ?P ) <==
%     input_policy_for_output ( ?input, ?policy, ?output_spec ) /\ {?output_spec.port ?P} /\
%     (
%     ({?policy.attribute.{name ?N; class ?T; value ?V}} /\ ~?output_spec.refinement.filter.{input ?input.port; name ?N; class ?T; value ?V; a Delete}) \/
%     ({?policy.attribute.{name ?N; class ?T’; value ?V’}} /\
%     ?output_spec.refinement.filter.{a Edit; input ?input.port; name ?N; class ?T’; value ?V’; new_class ?T; new_value ?V})
%     )
\begin{align*}
Ou&tputAttribute(n, t, v, p, attr) \Leftarrow \\
 &InputPolicyForOutput(input, policy, output) \land \\
 &hasPort(output, p) \land hasPort(input, port) \land ( \\
 &\begin{aligned}
     (&hasAttribute(policy, attr) \land hasName(attr, n) \land \\
      &hasClass(attr, t) \land hasValue(attr, v) \land \lnot \exists refi, filter. \\
      &hasRefinement(output, refi) \land hasFilter(refi, filter) \land \\
      &hasInput(filter, port) \land hasName(filter, n) \land \\
      &hasClass(filter, t) \land hasValue(filter, v)) \lor \\
     (&hasAttirbute(policy, attr) \land hasName(attr, n) \land \\
      &hasClass(attr, t') \land hasValue(attr, v') \land \\
      &hasRefinement(attr, refi) \land A(refi, Edit) \land \\
      &hasFilter(refi, filter) \land hasInput(filter, port) \land \\
      &hasName(filter, n) \land hasClass(filter, t') \land \\
      &hasValue(filter, v') \land hasNewClass(refi, t) \land \\
      &hasNewValue(refi, v))
 \end{aligned} \\
 &)
\end{align*}
Same as above, $OutputAttribute(n, t, v, p, attr)$ is a shorthand. However, different from them, it involves multiple nodes:
\begin{align*}
Ou&tputAttribute(n, t, v, p, attr) \equiv \\
 &\exists attr'. A(attr', Attribute) \land hasName(attr', n) \land \land \\
 &hasClass(attr', t) \land hasValue(attr', v) \land \\
 &\exists fl. A(fl, ForwardLink) \land hasOrigin(fl, attr) \land \\
 &hasPort(fl, p) \land hasRef(fl, attr')
\end{align*}
This means that we create a node $attr'$ which is of type $Attribute$, and assign its corresponding fields; we also create a linking relation (the node $fl$) between the original attribute $attr$ and the created attribute node $attr'$ at output $P$.

In addition to that, because we will often refer to the output attribute, we define this shorthand:
\begin{align*}
IO&Pair(in, out, p) \equiv \\
 &A(fl, ForwardLink) \land hasOrigin(fl, in) \land \\
 &hasPort(fl, p) \land hasRef(fl, out)
\end{align*}

\subsubsection{Output tag}

%
% OutputTagging( ?T, ?AR, ?VB, ?P ) <==
%     input_policy_for_output ( ?input, ?policy, ?output_spec) /\ { output_spec.port ?P } /\ {policy.tagging ?tagging} /\ {?tagging.{type ?T; attribute_ref ?AR; validity_binding ?VB}}
%     /\ ~deleted_attribute(?AR)
%     /\ ~(member(?X, ?VB) /\ deleted_attribute(?X))
% \begin{align*}
% &InputPolicyForOutput(input, policy, output) \land \\
% &hasPort(output, P) \land hasTagging(policy, tag) \land hasType(tag, T) \land \\
% &hasAttributeRef(tag, AR_0) \land \\
% &(ForwardLink(fl) \land hasOrigin(fl, AR_0) \land hasRef(fl, AR)) \land\\
% &(\forall vb. hasValidityBinding(tag, vb) \rightarrow \\
% &\quad \exists attr. ForwardLink(fl) \land hasOrigin(fl, vb) \land hasRef(fl, attr)) \\
% \rightarrow& OutputTagging(ot) \land hasType(ot, T) \land \\
% &hasAttributeRef(ot, AR) \land hasPort(ot, P) \land \\
% &(\forall vb. hasValidityBinding(tag, vb) \land ForwardLink(fl) \land \\
% &hasOrigin(fl, vb) \land hasRef(fl, attr) \\
% &\ \rightarrow hasValidityBinding(ot, attr))
% \end{align*}
\begin{align*}
Ou&tputTag(t, ar, p, tag) \Leftarrow \\
 &InputPolicyForOutput(input, policy, output) \land \\
 &hasPort(output, p) \land hasTag(policy, tag) \land hasCategory(tag, t) \land \\
 &hasAttributeRef(tag, ar_0) \land IOPair(ar_0, ar, p) \\
 % &hasAttributeRef(tag, ar_0) \land \\
 % &(A(fl, ForwardLink) \land hasOrigin(fl, ar_0) \land hasRef(fl, ar)) \land\\
 &(\forall vb. hasValidityBinding(tag, vb) \rightarrow \\
 &\quad \exists attr. IOPair(vb, attr, p))
 % &\quad \exists attr. A(fl, ForwardLink) \land hasOrigin(fl, vb) \land hasRef(fl, attr))
\end{align*}
where the full form of $OutputTag$ is:
\begin{align*}
Ou&tputTag(t, ar, p, tag) \equiv \\
 &\exists ot. A(ot, Tag) \land hasCategory(ot, t) \land \\
 &hasAttributeRef(ot, ar) \land hasPort(ot, p) \land \\
 &(\forall vb,attr. hasValidityBinding(tag, vb) \land IOPair(vb, attr, p) \\
 % &\forall vb,fl,attr. (hasValidityBinding(tag, vb) \land A(fl, ForwardLink) \land \\
 % &\ hasOrigin(fl, vb) \land hasRef(fl, attr) \\
 &\ \rightarrow hasValidityBinding(ot, attr))
\end{align*}

\subsubsection{Output prohibition}
% OutputProhibition( ?M, ?VB, ?AC, ?P ) <==
%     input_policy_for_output ( ?input, ?policy, ?output_spec) /\ { output_spec.port ?P } /\ {policy.prohibition ?pr} /\ {?pr.{use_mode ?M; validity_binding ?VB; activation_condition ?AC} /\ ~(member(?X, ?VB) /\ deleted_attribute(?X))
\begin{align*}
Ou&tputProhibition( m, ac, p, pr ) \Leftarrow \\
 &InputPolicyForOutput(input, policy, output) \land \\
 &hasPort(output, p) \land hasProhibition(policy, pr) \land \\
 &hasMode(pr, m) \land hasActivationCondition(pr, ac) \land \\
 &\forall vb. (hasValidityBinding(pr, vb) \rightarrow \exists attr. IOPair(vb, attr, p))
\end{align*}
where the full form of $OutputProhibition$ is:
\begin{align*}
Ou&tputProhibition( m, ac, p, pr ) \equiv \\
 &\exists op. A(op, Prohibition) \land hasMode(op, m) \land \\
 &hasActivationCondition(op, ac) \land hasPort(op, p) \land \\
 &\forall vb,attr. (hasValidityBinding(pr, vb) \land IOPair(vb, attr, p) \\
 &\ \rightarrow hasValidityBinding(op, attr))
\end{align*}

\subsubsection{Output obligation}
% OutputObligation(  ?OC, ?Arg, ?VB, ?AC, ?P ) <==
%     input_policy_for_output ( ?input, ?policy, ?output_spec) /\ { output_spec.port ?P } /\ {policy.obligation ?ob} /\ {?ob.{obligation_class ?OC; argument ?Arg; validity_binding ?VB; activation_condition ?AC}} /\ ~(member(?X, ?VB) /\ deleted_attribute(?X))
\begin{align*}
Ou&tputObligation( oc, args, ac, p, ob ) \Leftarrow \\
 &InputPolicyForOutput(input, policy, output) \land \\
 &hasPort(output, p) \land hasObligation(policy, ob) \land \\
 &hasObligationClass(ob, oc) \land hasArgument(ob, args) \land \\
 &\forall x. (member(args, x) \rightarrow \exists x'. IOPair(x, x', p)) \land \\
 &hasActivationCondition(ob, ac) \land \\
 &\forall vb. (hasValidityBinding(ob, vb) \rightarrow \exists attr. IOPair(vb, attr, p))
\end{align*}
where the full form of $OutputObligation$ is:
\begin{align*}
Ou&tputObligation( oc, args, ac, p, ob) \equiv \\
 &\exists oo. A(oo, Obligation) \land hasObligationClass(oo, oc) \land \\
 &\exists args'. hasArgument(oo, args') \land \\
 &\ (\forall x. member(args, x) \rightarrow member(args', x)) \land \\
 &(\forall vb, attr. hasValidityBinding(ob, vb) \land IOPair(vb, attr, p) \\
 &\ \rightarrow hasValidityBinding(oo, attr))
\end{align*}
where $member(args, x)$ means $x$ is a member of the list (RDF Collection) $args$.

\end{document}